\documentclass[10pt,journal,compsoc]{IEEEtran}

\ifCLASSOPTIONcompsoc

  \usepackage[nocompress]{cite}
\else
  \usepackage{cite}
\fi

\ifCLASSINFOpdf
  \usepackage[pdftex]{graphicx}
  \DeclareGraphicsExtensions{.pdf,.jpeg,.png}
    \DeclareGraphicsExtensions{.eps}
\usepackage{epstopdf}
\else
\fi

\usepackage{amsmath}

\usepackage{array}

\ifCLASSOPTIONcompsoc
 \usepackage[caption=false,font=footnotesize,labelfont=sf,textfont=sf]{subfig}
\else
 \usepackage[caption=false,font=footnotesize]{subfig}
\fi
\usepackage{booktabs} 
\usepackage{amssymb}

\usepackage{xcolor}
\usepackage[breaklinks=true,bookmarks=false]{hyperref}
\hypersetup{
    colorlinks,
    linkcolor={red!90!black},
    citecolor={green!90!black},
    urlcolor={blue!90!black}
}
\usepackage{multirow}

\begin{document}

\title{Look into Person: Joint Body Parsing \& Pose Estimation Network and A New Benchmark}

\author{Xiaodan~Liang, Ke~Gong, Xiaohui~Shen, and~Liang~Lin
\IEEEcompsocitemizethanks{\IEEEcompsocthanksitem X. Liang, K. Gong and L. Lin are with the School of Data and Computer Science,
Sun Yat-sen University, China.\protect
\IEEEcompsocthanksitem X. Shen is with Adobe Research.}
\thanks{X. Liang and K. Gong contribute equally to this paper. Corresponding author: Liang Lin (E-mail: linliang@ieee.org)}}

\IEEEtitleabstractindextext{%
\begin{abstract}
Human parsing and pose estimation have recently received considerable interest due to their substantial application potentials. However, the existing datasets have limited numbers of images and annotations and lack a variety of human appearances and coverage of challenging cases in unconstrained environments. In this paper, we introduce a new benchmark named ``Look into Person (LIP)" that provides a significant advancement in terms of scalability, diversity, and difficulty, which are crucial for future developments in human-centric analysis. This comprehensive dataset contains over 50,000 elaborately annotated images with 19 semantic part labels and 16 body joints, which are captured from a broad range of viewpoints, occlusions, and background complexities. Using these rich annotations, we perform detailed analyses of the leading human parsing and pose estimation approaches, thereby obtaining insights into the successes and failures of these methods. To further explore and take advantage of the semantic correlation of these two tasks, we propose a novel joint human parsing and pose estimation network to explore efficient context modeling, which can simultaneously predict parsing and pose with extremely high quality. Furthermore, we simplify the network to solve human parsing by exploring a novel self-supervised structure-sensitive learning approach, which imposes human pose structures into the parsing results without resorting to extra supervision. The datasets, code and models are available at \url{http://www.sysu-hcp.net/lip/}.
\end{abstract}

}

\maketitle

\IEEEdisplaynontitleabstractindextext

\IEEEpeerreviewmaketitle

\ifCLASSOPTIONcompsoc
\IEEEraisesectionheading{\section{Introduction}\label{sec:introduction}}
\else
\section{Introduction}
\label{sec:introduction}
\fi

\IEEEPARstart{C}{omprehensive} human visual understanding of scenarios in the wild, which is regarded as one of the most fundamental problems in computer vision, could have a crucial impact in many higher-level application domains, such as person re-identification~\cite{zhao2013unsupervised}, video surveillance~\cite{wang2014deformable}, human behavior analysis~\cite{gan2016concepts,liang2015proposal} and automatic product recommendation~\cite{kalantidis2013getting}. Human parsing {(also named semantic part segmentation)} aims to segment a human image into multiple parts with fine-grained semantics (e.g., body parts and clothing) and provides a more detailed understanding of image contents, whereas human pose estimation focuses on determining the precise locations of important body joints. Human parsing and pose estimation are two critical and correlated tasks in analyzing images of humans by providing both pixel-wise understanding and high-level joint structures.

\begin{figure*}[!t]
\centering
\subfloat[MPII and LSP]{\includegraphics[width=1\linewidth]{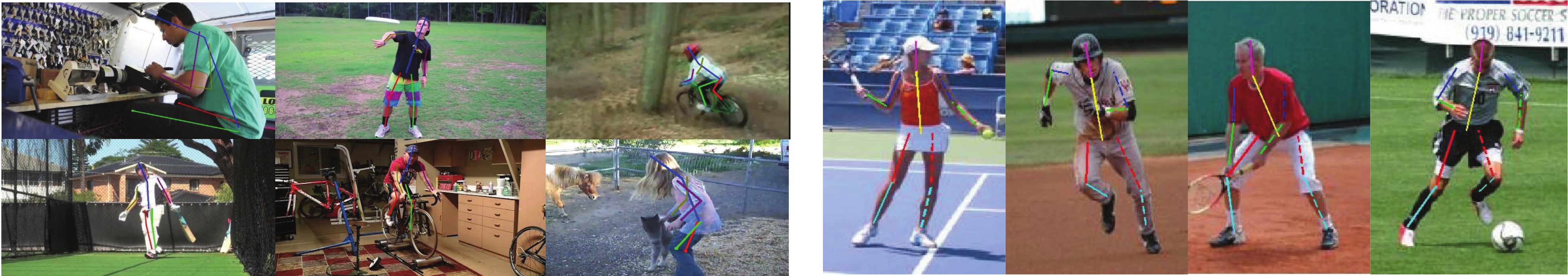}}
\hfil
\vspace{-2mm}
\subfloat[ATR and PASCAL-Person-Part]{\includegraphics[width=1\linewidth]{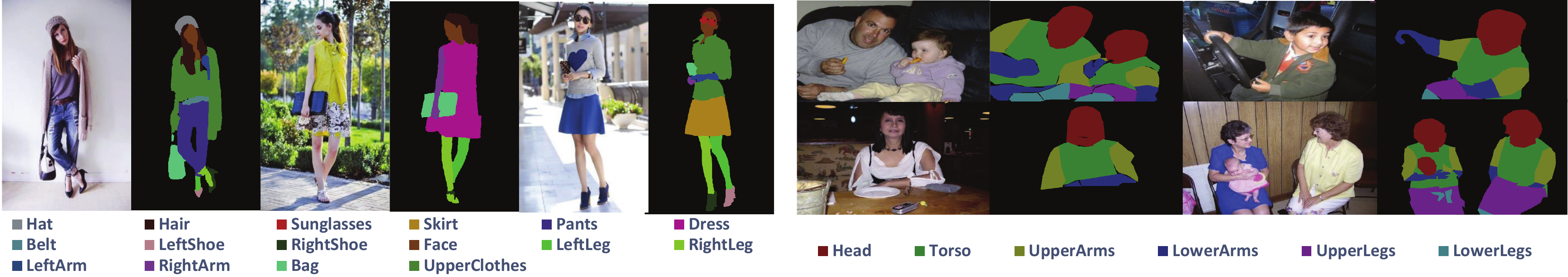}}
\hfil
\vspace{-2mm}
\subfloat[LIP]{\includegraphics[width=1\linewidth]{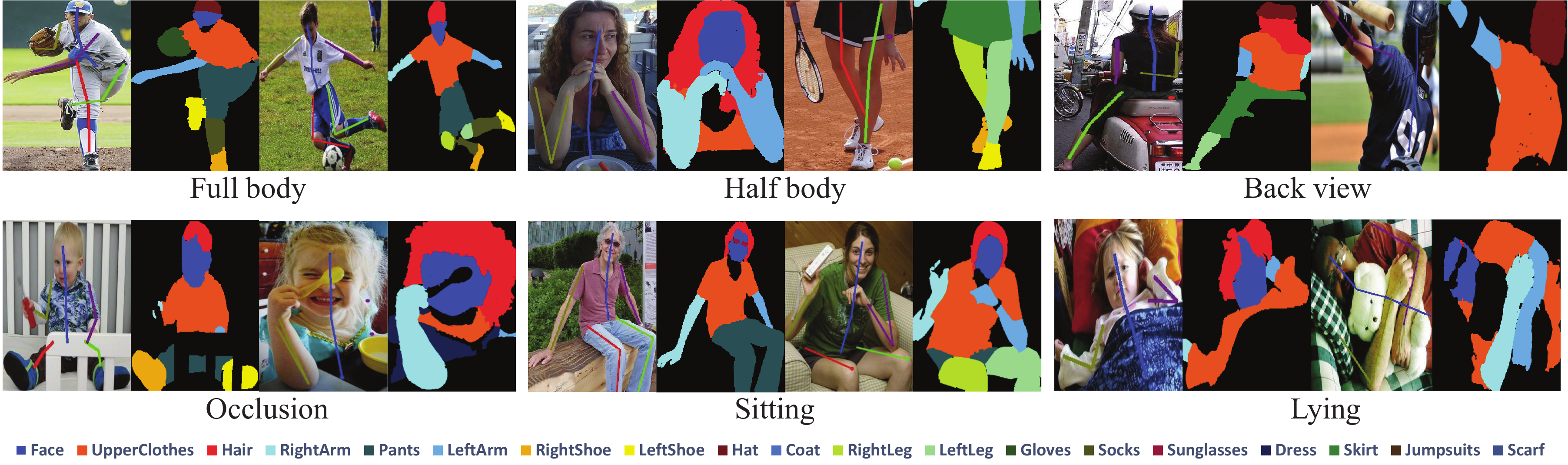}}
\vspace{-2mm}
\caption{Annotation examples for our ``Look into Person (LIP)'' dataset and existing datasets. (a) The images in the MPII dataset~\cite{andriluka14cvpr} (left) and LSP dataset~\cite{Johnson10} (right) with only body joint annotations. (b) The images in the ATR dataset~\cite{Co-CNN} (left)  are fixed in size and only contain stand-up person instances in the outdoors. The images in the PASCAL-Person-Part dataset~\cite{chen2014detect} (right) also have lower scalability and only contain 6 coarse labels. (c) The images in our LIP dataset have high appearance variability and complexity, and they are annotated with both human parsing and pose labels.}
\vspace{-2mm}
\label{fig:dataset_example}
\end{figure*}

Recently, convolutional neural networks (CNNs) have achieved exciting success in human parsing~\cite{ATR,Co-CNN,liang2015semantic} and pose estimation~\cite{newell2016stacked,Wei_2016_CVPR}. Nevertheless, as demonstrated in many other problems, such as object detection~\cite{liang2015towards} and semantic segmentation~\cite{crfasrnn}, the performance of such CNN-based approaches heavily relies on the availability of annotated images for training. To train a human parsing or pose network with potential practical value in real-world applications, it is highly desired to have a large-scale dataset that is composed of representative instances with varied clothing appearances, strong articulation, partial (self-)occlusions, truncation at image borders, diverse viewpoints and background clutters. Although training sets exist for special scenarios, such as fashion pictures~\cite{Yamaguchiparsing13,Dongparsing13,ATR,Co-CNN} and people in constrained situations (e.g., upright)~\cite{chen2014detect}, these datasets are limited in their coverage and scalability, as shown in Fig.~\ref{fig:dataset_example}. The largest public human parsing dataset~\cite{Co-CNN} to date only contains 17,000 fashion images, while others only include thousands of images. The MPII Human Pose dataset~\cite{andriluka14cvpr} is the most popular benchmark for evaluating articulated human pose estimation methods, and this dataset includes approximately 25K images that contain over 40K people with annotated body joints. However, all these datasets only focus on addressing different aspects of human analysis by defining discrepant annotation sets. There are no available unified datasets with both human parsing and pose annotations for holistic human understanding, until our work fills this gap.

\begin{figure}[t]
\centering
   \includegraphics[width=1.0\linewidth]{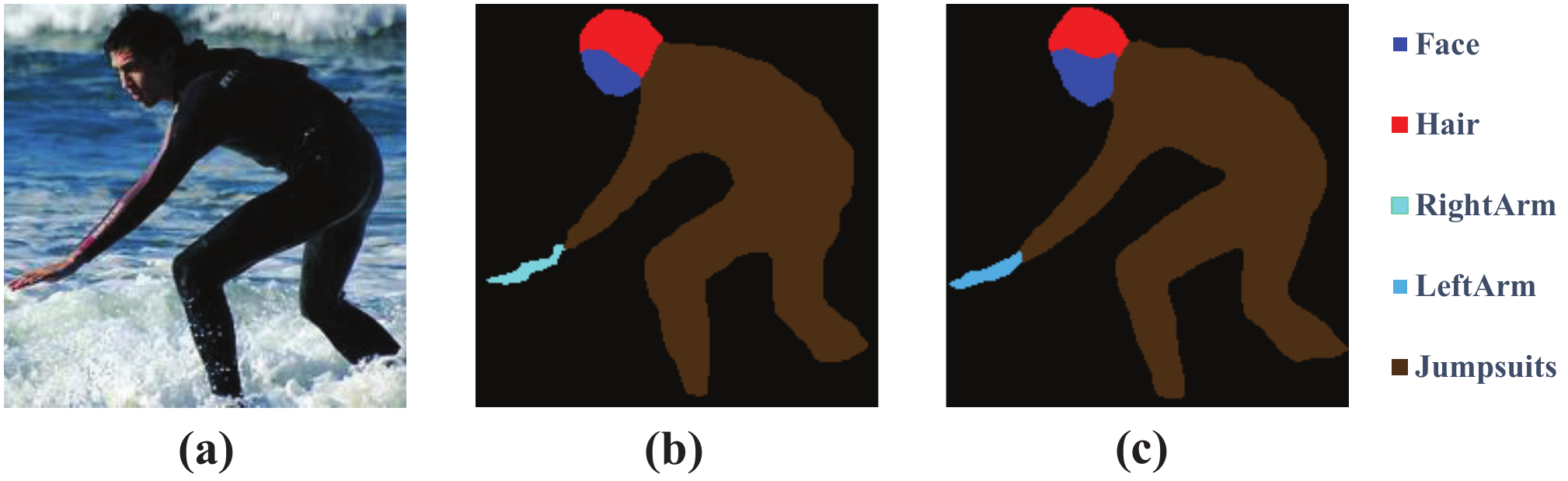}
\vspace{-8mm}
\caption{This example shows that human body structural information is helpful for human parsing. (a) The original image. (b) The parsing results by attention-to-scale~\cite{chen2015attention}, where the left arm is incorrectly labeled as the right arm. (c) Our parsing results successfully incorporate the structure information to generate reasonable outputs.}
\vspace{-6mm}
\label{fig: example_compare}
\end{figure}

Furthermore, to the best of our knowledge, no attempt has been made to establish a standard representative benchmark that aims to cover a wide range of challenges for the two human-centric tasks. The existing datasets do not provide an evaluation server with a secret test set to avoid potential dataset over-fitting, which hinders further development on this topic. With the new benchmark named "Look into Person (LIP)", we provide a public server for automatically reporting evaluation results. Our benchmark significantly advances the state-of-the-art in terms of appearance variability and complexity, and it includes 50,462 human images with pixel-wise annotations of 19 semantic parts and 16 body joints.

The recent progress in human parsing~\cite{chen2015attention,xia2015zoom,LinASPL,Yamaguchiparsing13,Dongparsing13,SimoSerraACCV2014,M-CNN,Co-CNN} has been achieved by improving the feature representations using CNNs and recurrent neural networks. To capture rich structure information, these approaches combine CNNs and graphical models (e.g., conditional random fields (CRFs)), similar to the general object segmentation approaches~\cite{crfasrnn,chen2014semantic,wang2015joint}. However, when evaluated on the new LIP dataset, the results of some existing methods~\cite{badrinarayanan2015segnet,long2014fully,chen2014semantic,chen2015attention} are unsatisfactory. Without imposing human body structure priors, these general approaches based on bottom-up appearance information occasionally tend to produce unreasonable results (e.g., right arm connected with left shoulder), as shown in Fig.~\ref{fig: example_compare}. Human body structural information has previously been well explored in human pose estimation~\cite{yang2016end,Chen_NIPS14}, where dense joint annotations are provided. However, since human parsing requires more extensive and detailed predictions than pose estimation, it is difficult to directly utilize joint-based pose estimation models in pixel-wise predictions to incorporate the complex structure constraints. We demonstrate that the human joint structure can facilitate the pixel-wise parsing prediction by incorporating higher-level semantic correlations between human parts.

For pose estimation, increasing research efforts~\cite{Chu_2017_CVPR,bulat2016human,dantone2013human,park2017attribute} have been devoted to learning the relationships between human body parts and joints. Some studies~\cite{Chu_2017_CVPR,bulat2016human} explored encoding part constraints and contextual information for guiding the network to focus on informative regions (human parts) to predict more precise locations of the body joints, which achieved state-of-the-art performance. Conversely, some pose-guided human parsing methods~\cite{dong2014towards,xia2016pose} also sufficiently utilized the peculiarity and relationships of these two correlated tasks. However, previous works generally solve these two problems separately or alternatively.

\begin{table*}[t]
\centering
\normalsize
\caption{Overview of the publicly available datasets for human parsing and pose estimation. For each dataset, we report the number of images in the training, validation and test sets; parsing categories, including background; and body joints.}
\vspace{-3mm}
\label{tab:dataset_num}
\begin{tabular}{ccccccc}
\toprule[0.7pt]
    Dataset                                  & \#Total & \#Training   & \#Validation  & \#Test & Parsing Categories & Body Joints   \\ \hline 
    Fashionista~\cite{yamaguchi2012parsing} (Parsing)  &  685     & 456          & -           & 229       &     56          &      -        \\
    PASCAL-Person-Part~\cite{chen2014detect} (Parsing) &  3533    & 1,716        & -           & 1,817     &     7           &      -        \\
    ATR~\cite{Co-CNN} (Parsing)                        &  17,700  & 16,000       & 700         & 1,000     &     18          &      -        \\ 
    LSP~\cite{Johnson10} (Pose)                      &  2,000     & 1,000        & -           & 1,000     &     -           &      14       \\
    MPII~\cite{andriluka14cvpr} (Pose)               &  24987     & 18079        & -           & 6908      &     -           &      16  \\ 
    {J-HMDB~\cite{Jhuang:ICCV:2013} (Pose)} &  31838     & -        & -           & -      &     2           &      13  \\ \hline 
    LIP                                              &  {50,462}  & 30,462       & 10,000    & 10,000    &  20      &      16       \\ 
\toprule[0.7pt]
\vspace{-2mm}
\end{tabular}
\end{table*}

In this work, we aim to seamlessly integrate human parsing and pose estimation under a unified framework. We use a shared deep residual network for feature extraction, after which there are two distinct small networks to encode and predict the contextual information and results. Then, a simple yet efficient refinement network tailored for both parsing and pose prediction is proposed to explore efficient context modeling, which makes human parsing and pose estimation mutually beneficial. In our unified framework, we propose a scheme to incorporate multi-scale feature combinations and iterative location refinement together, which are often posed as two different coarse-to-fine strategies that are widely investigated for human parsing and pose estimation separately. To highlight the merit of unifying the two highly correlated and complementary tasks within an end-to-end framework, we name our framework the joint human parsing and pose estimation network (JPPNet).

However, note that annotating both pixel-wise labeling maps and pose joints is unrealized in previous human-centric datasets. Therefore, in this work, we also design a simplified network suited to general human parsing datasets and networks with no need for pose annotations. To explicitly enforce the produced parsing results to be semantically consistent with the human pose and joint structures, we propose a novel structure-sensitive learning approach for human parsing. In addition to using the traditional pixel-wise part annotations as the supervision, we introduce a structure-sensitive loss to evaluate the quality of the predicted parsing results from a joint structure perspective. This means that a satisfactory parsing result should be able to preserve a reasonable joint structure (e.g., the spatial layouts of human parts). We generate approximated human joints directly from the parsing annotations and use them as the supervision signal for the structure-sensitive loss. This self-supervised structure-sensitive network is a simplified verson of our JPPNet, denoted as SS-JPPNet, which is appropriate for the general human parsing datasets without pose annotations

Our contributions are summarized in the following three aspects. 1) We propose a new large-scale benchmark and an evaluation server to advance the human parsing and pose estimation research, in which 50,462 images with pixel-wise annotations on 19 semantic part labels and 16 body joints are provided. 2) By experimenting on our benchmark, we present detailed analyses of the existing human parsing and pose estimation approaches to obtain some insights into the successes and failures of these approaches and thoroughly explore the relationship between the two human-centric tasks. 3) We propose a novel joint human parsing and pose estimation network, which incorporates the multi-scale feature connections and iterative location refinement in an end-to-end framework to investigate efficient context modeling and then enable parsing and pose tasks that are mutually beneficial to each other. This unified framework achieves state-of-the-art performance for both human parsing and pose estimation tasks. The simplified network for human parsing task with self-supervised structure-sensitive learning also significantly surpasses the previous methods on both the existing PASCAL-Person-Part dataset~\cite{chen2014detect} and our new LIP dataset.

\section{Related Work}

\textbf{Human parsing and pose datasets:}
The commonly used publicly available datasets for human parsing and pose are summarized in Table~\ref{tab:dataset_num}. For human parsing, the previous datasets were labeled with a limited number of images or categories. The largest dataset~\cite{Co-CNN} to date only contains 17,000 fashion images with mostly upright fashion models. These small datasets are unsuitable for training models with complex appearance representations and multiple components~\cite{johnson2011learning,taskar2013modec,dantone2013human}, which could perform better. For human pose, the LSP dataset~\cite{Johnson10} only contains sports people, and it fails to cover real-life challenges. The MPII dataset~\cite{andriluka14cvpr} has more images and a wider coverage of activities that cover real-life challenges, such as truncation, occlusions, and variability of imaging conditions. However, this dataset only provides 2D pose annotations. {J-HMDB~\cite{Jhuang:ICCV:2013} provides densely annotated image sequences and a larger number of videos for 21 human actions. Although the puppet mask and human pose are annotated in the all 31838 frames, detailed part segmentations are not labeled. }

Our proposed LIP benchmark dataset is the first effort that focuses on the two human-centric tasks. Containing 50,462 images annotated with 20 parsing categories and 16 body joints, our LIP dataset is the largest and most comprehensive human parsing and pose dataset to date. Some other datasets in the vision community were dedicated to the tasks of clothes recognition, retrieval~\cite{liuLQWTcvpr16DeepFashion} and fashion modeling~\cite{simo2015neuroaesthetics}, whereas our LIP dataset only focuses on human parsing and pose estimation.

\textbf{Human parsing approaches:}  
Recently, many research efforts have been devoted to human parsing~\cite{Co-CNN,yamaguchi2012parsing,Yamaguchiparsing13,SimoSerraACCV2014,M-CNN,xia2015zoom,chen2015attention}. For example, Liang et al.~\cite{Co-CNN} proposed a novel Co-CNN architecture that integrates multiple levels of image contexts into a unified network. In addition to human parsing, there has also been increasing research interest in the part segmentation of other objects, such as animals or cars~\cite{wang2014semantic,wang2015joint,lu2014parsing}. To capture the rich structure information based on the advanced CNN architecture, common solutions include the combination of CNNs and CRFs~\cite{chen2016deeplab,crfasrnn} and adopting multi-scale feature representations~\cite{chen2016deeplab,chen2015attention,xia2015zoom}. Chen et al.~\cite{chen2015attention} proposed an attention mechanism that learns to weight the multi-scale features at each pixel location. 

\textbf{Pose estimation approaches: }
Articulated human poses are generally modeled using a combination of a unary term and pictorial structures~\cite{andriluka2009pictorial} or graph model, e.g., mixture of body parts~\cite{YangR_CVPR_2011,Chen_NIPS14,ouyang2014multi}. With the introduction of DeepPose~\cite{deeppose13}, which formulates the pose estimation problem as a regression problem using a standard convolutional architecture, research on human pose estimation began to shift from classic approaches to deep networks. For example, Wei et al.~\cite{Wei_2016_CVPR} incorporated the inference of the spatial correlations among body parts within ConvNets. Newell et al. proposed a stacked hourglass network~\cite{newell2016stacked} using a repeated pooling down and upsampling process to learn the spatial distribution.

Some previous works~\cite{dong2014towards,xia2016pose} explored human pose information to guide human parsing by generating ``pose-guided'' part segment proposals. Additionally, some works~\cite{bulat2016human,Chu_2017_CVPR} generated attention maps of the body part to guide pose estimation. To further utilize the advantages of parsing and pose and their relationships, our focus is a joint human parsing and pose estimation network, which can simultaneously predict parsing and pose with extremely high quality. Additionally, to leverage the human joint structure more effortlessly and efficiently, we simplify the network and propose a self-supervised structure-sensitive learning approach.

{The rest of this paper is organized as follows. We present the analysis of existing human parsing and pose estimation datasets and then introduce our new LIP benchmark in Section 3. In Section 4, we present the empirical study of current methods based on our LIP benchmark and discuss the limitations of these methods. Then, to address the challenges raised by LIP, we propose a unified framework for simultaneous human parsing and pose estimation in Section 5. At last, more detailed comparisons between our approach and state-of-the-art methods are exhibited in Section 6.}

\begin{figure}[t]
\centering
   \includegraphics[width=1.0\linewidth]{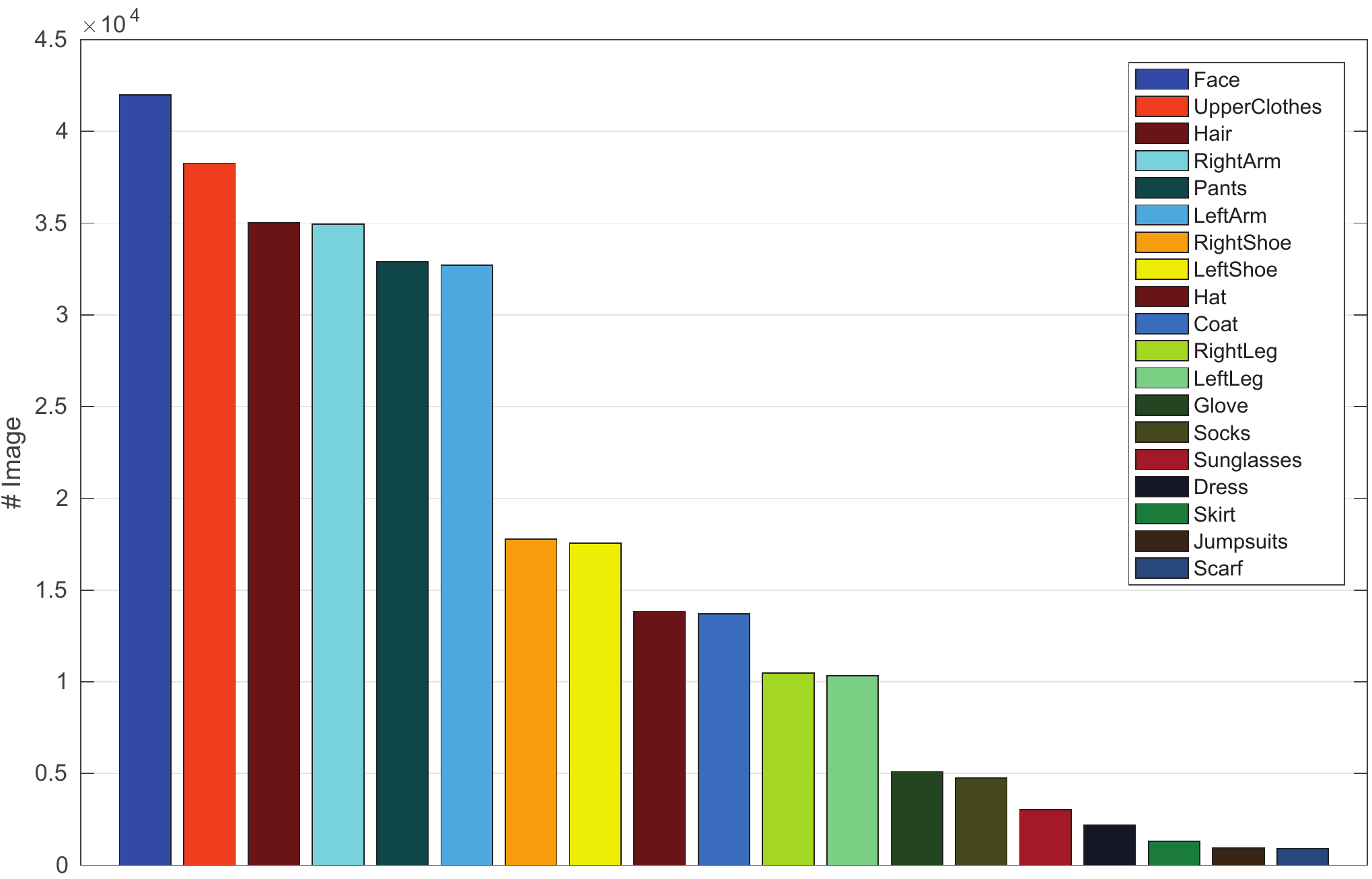}
\vspace{-8mm}
\caption{The data distribution on 19 semantic part labels in the LIP dataset.}
\vspace{-4mm}
\label{fig: dataset_analysis_label}
\end{figure}

\section{Look into Person Benchmark}
In this section, we introduce our new ``Look into Person (LIP)'' dataset, which is a new large-scale dataset that focuses on semantic understanding of human bodies and that has several appealing properties. First, with 50,462 annotated images, LIP is an order of magnitude larger and more challenging than previous similar datasets~\cite{yamaguchi2012parsing,chen2014detect,Co-CNN}. Second, LIP is annotated with elaborate pixel-wise annotations with 19 semantic human part labels and one background label for human parsing and 16 body joint locations for pose estimation. Third, the images collected from the real-world scenarios contain people appearing with challenging poses and viewpoints, heavy occlusions, various appearances and in a wide range of resolutions. Furthermore, the background of the images in the LIP dataset is also more complex and diverse than that in previous datasets. Some examples are shown in Fig.~\ref{fig:dataset_example}. With the LIP dataset, we propose a new benchmark suite for human parsing and pose estimation together with a standard evaluation server where the test set will be kept secret to avoid overfitting.

\begin{figure}[t]
\centering
   \includegraphics[width=1.0\linewidth]{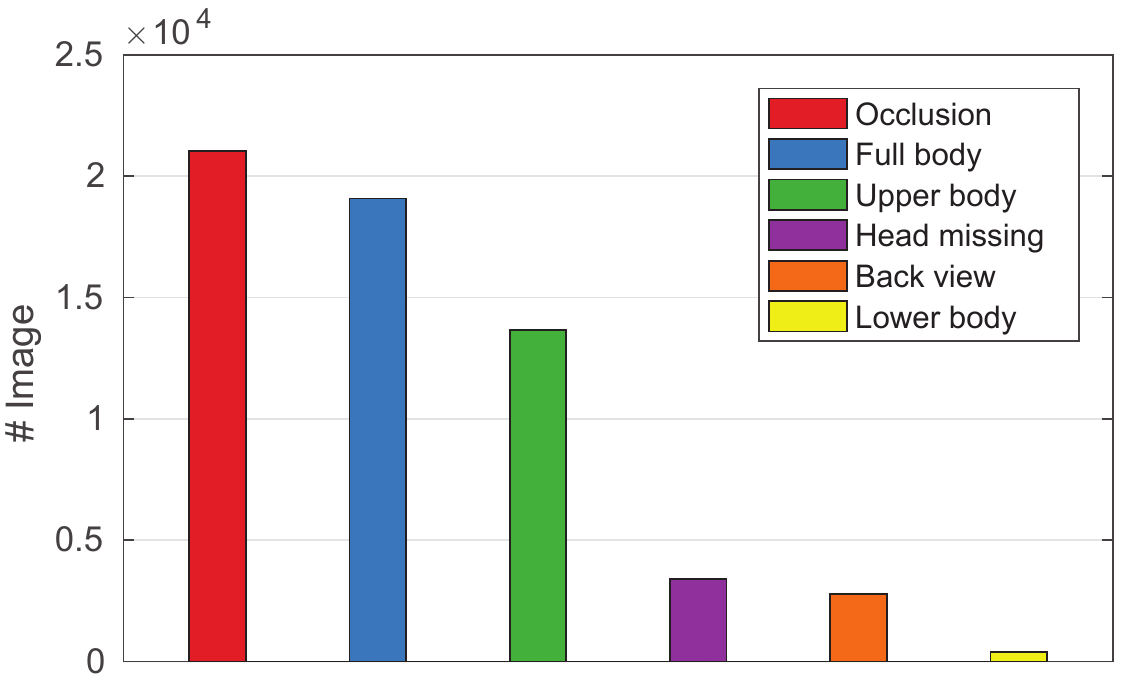}
\vspace{-8mm}
\caption{The numbers of images that show diverse visibilities in the LIP dataset, including occlusion, full body, upper body, lower body, head missing and back view.}
\vspace{-4mm}
\label{fig: dataset_analysis_factor}
\end{figure}

\subsection{Image Annotation}
The images in the LIP dataset are cropped person instances from Microsoft COCO~\cite{DBLP:journals/corr/LinMBHPRDZ14} training and validation sets. We defined 19 human parts or clothes labels for annotation, which are hat, hair, sunglasses, upper clothes, dress, coat, socks, pants, gloves, scarf, skirt, jumpsuit, face, right arm, left arm, right leg, left leg, right shoe, and left shoe, as well as a background label. Similarly, we provide rich annotations for human poses, where the positions and visibility of 16 main body joints are annotated. Following~\cite{andriluka14cvpr}, we annotate joints in a ``person-centric'' manner, which means that the left/right joints refer to the left/right limbs of the person. At test time, this requires pose estimation with both a correct localization of the limbs of a person along with the correct match to the left/right limb.

{We implemented an annotation tool and generate multi-scale superpixels of images to speed up the annotation. More than 100 students are trained well to accomplish annotation work which lasts for five months. We supervise the whole annotation process and check the results periodically to control the annotation quality. Finally, we conduct a second-round check for each annotated image and selecte 50,000 usable and well-annotated images strictly and carefully from over 60,000 submitted images. }

\subsection{Dataset Splits}
In total, there are 50,462 images in the LIP dataset, including 19,081 full-body images, 13,672 upper-body images, 403 lower-body images, 3,386 head-missing images, 2,778 back-view images and 21,028 images with occlusions. We split the images into separate training, validation and test sets. Following random selection, we arrive at a unique split that consists of 30,462 training and 10,000 validation images with publicly available annotations, as well as 10,000 test images with annotations withheld for benchmarking purposes.

Furthermore, to stimulate the multiple-human parsing research, we collect the images with multiple person instances in the LIP dataset to establish the first standard and comprehensive benchmark for multiple-human parsing and pose estimation. Our LIP multiple-human parsing and pose dataset contains 4,192 training, 497 validation and 458 test images, in which there are 5,147 multiple-person images in total.
\subsection{Dataset Statistics}
In this section, we analyze the images and categories in the LIP dataset in detail. In general, face, arms, and legs are the most remarkable parts of a human body. However, human parsing aims to analyze every detailed region of a person, including different body parts and different categories of clothes. We therefore define 6 body parts and 13 clothes categories. Among these 6 body parts, we divide arms and legs into the left side and right side for a more precise analysis, which also increases the difficulty of the task. For clothes classes, we  not only have common clothes such as upper clothes, pants, and shoes but also have infrequent categories, such as skirts and jumpsuits. Furthermore, small-scale accessories such as sunglasses, gloves, and socks are also taken into account. The numbers of images for each semantic part label are presented in Fig.~\ref{fig: dataset_analysis_label}.

In contrast to other human image datasets, the images in the LIP dataset contain diverse human appearances, viewpoints, and occlusions. Additionally, more than half of the images suffer from occlusions of different degrees. An occlusion is considered to have occurred if any of the semantic parts or body joints appear in the image but are occluded or invisible. In more challenging cases, the images contain person instances in a back view, which gives rise to more ambiguity in the left and right spatial layouts. The numbers of images of different appearances (i.e., occlusion, full body, upper body, head missing, back view and lower body) are summarized in Fig.~\ref{fig: dataset_analysis_factor}.

\begin{table}[t]
\centering
\scriptsize
\caption{Comparison of human parsing performance with five state-of-the-art methods on the LIP validation set.}
\vspace{-3mm}
\label{tab: lip_val}
\begin{tabular}{cccc}
\toprule[0.7pt]
   Method                                      & Overall accuracy   & Mean accuracy    & Mean IoU    \\ \hline 
   SegNet~\cite{badrinarayanan2015segnet}      & 69.04              & 24.00            & 18.17       \\
   FCN-8s~\cite{long2014fully}                 & 76.06              & 36.75            & 28.29       \\
   DeepLab (VGG-16)~\cite{chen2016deeplab}     & 82.66              & 51.64            & 41.64       \\
   Attention~\cite{chen2015attention}          & 83.43              & 54.39            & 42.92       \\ 
   DeepLab (ResNet-101)~\cite{chen2016deeplab} & 84.09              & 55.62            & 44.80       \\    \hline
   JPPNet {(with pose info)}          & \textbf{86.39}     & \textbf{62.32}   & \textbf{51.37}       \\
   SS-JPPNet                                   & 84.36              & 54.94            & 44.73       \\
\toprule[0.7pt]
\vspace{-6mm}
\end{tabular}
\end{table}

\begin{table}[t]
\centering
\scriptsize
\caption{Comparison of human parsing performance with five state-of-the-art methods on the LIP test set.}
\vspace{-3mm}
\label{tab: lip_test}
\begin{tabular}{cccc}
\toprule[0.7pt]
   Method                                       & Overall accuracy & Mean accuracy   & Mean IoU  \\ \hline 
   SegNet~\cite{badrinarayanan2015segnet}       & 69.10            & 24.26           & 18.37    \\
   FCN-8s~\cite{long2014fully}                  & 76.28            & 37.18           & 28.69    \\
   DeepLab (VGG-16)~\cite{chen2016deeplab}      & 82.89            & 51.53           & 41.56    \\
   Attention~\cite{chen2015attention}           & 83.56            & 54.28           & 42.97    \\ 
   DeepLab (ResNet-101)~\cite{chen2016deeplab}  & 84.25            & 55.64           & 44.96    \\    \hline
   JPPNet {(with pose info)}           & \textbf{86.48}   & \textbf{62.25}  & \textbf{51.36}       \\
   SS-JPPNet                                    & 84.53            & 54.81           & 44.59    \\
\toprule[0.7pt]
\vspace{-4mm}
\end{tabular}
\end{table}

\begin{figure}[t]
\centering
   \includegraphics[width=1.0\linewidth]{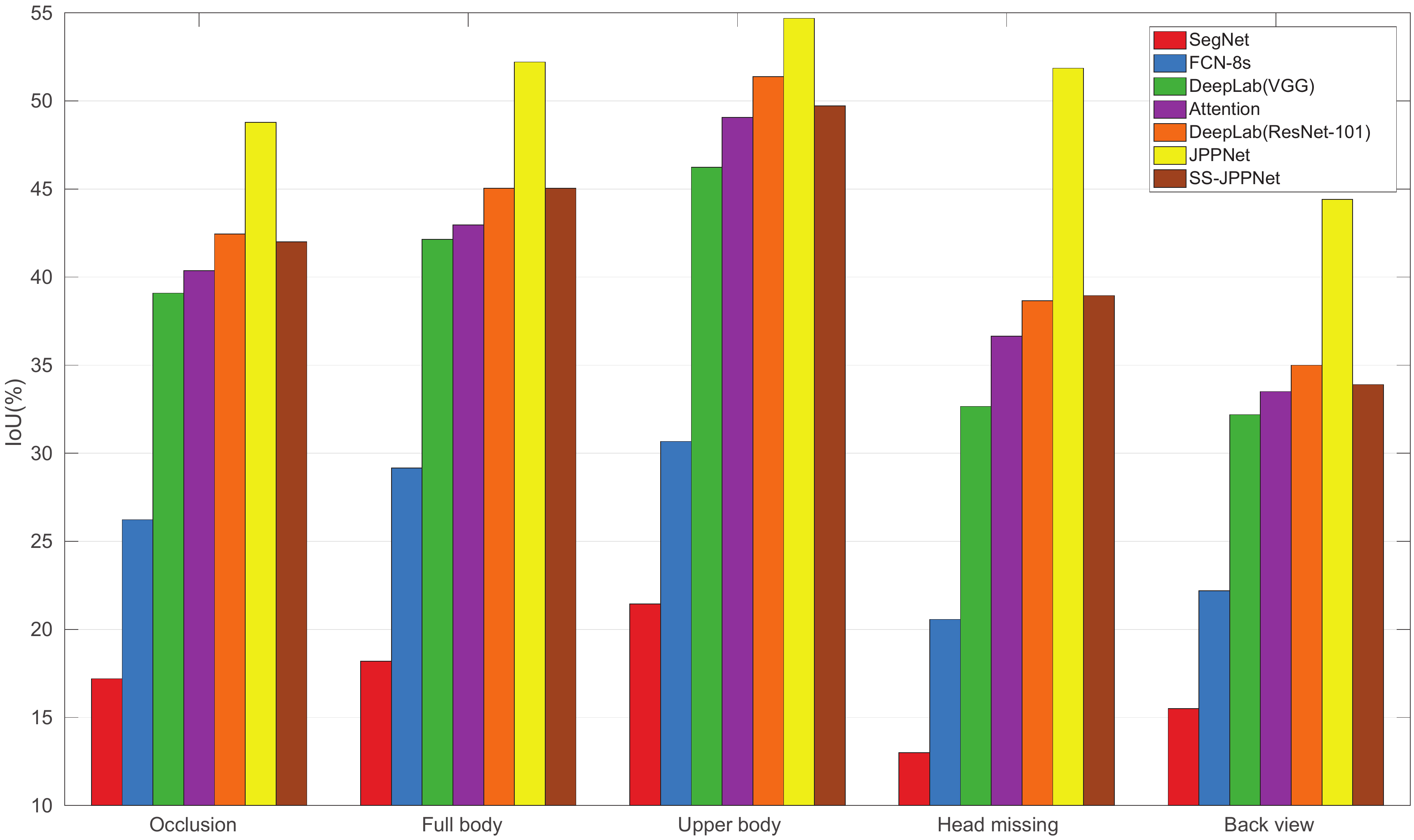}
\vspace{-8mm}
\caption{Human parsing performance comparison evaluated on the LIP validation set with different appearances, including occlusion, full body, upper body, head missing and back view.}
\vspace{-2mm}
\label{fig: analysis_val}
\end{figure}

\begin{figure*}[t]
\centering
\includegraphics[width=0.9\linewidth]{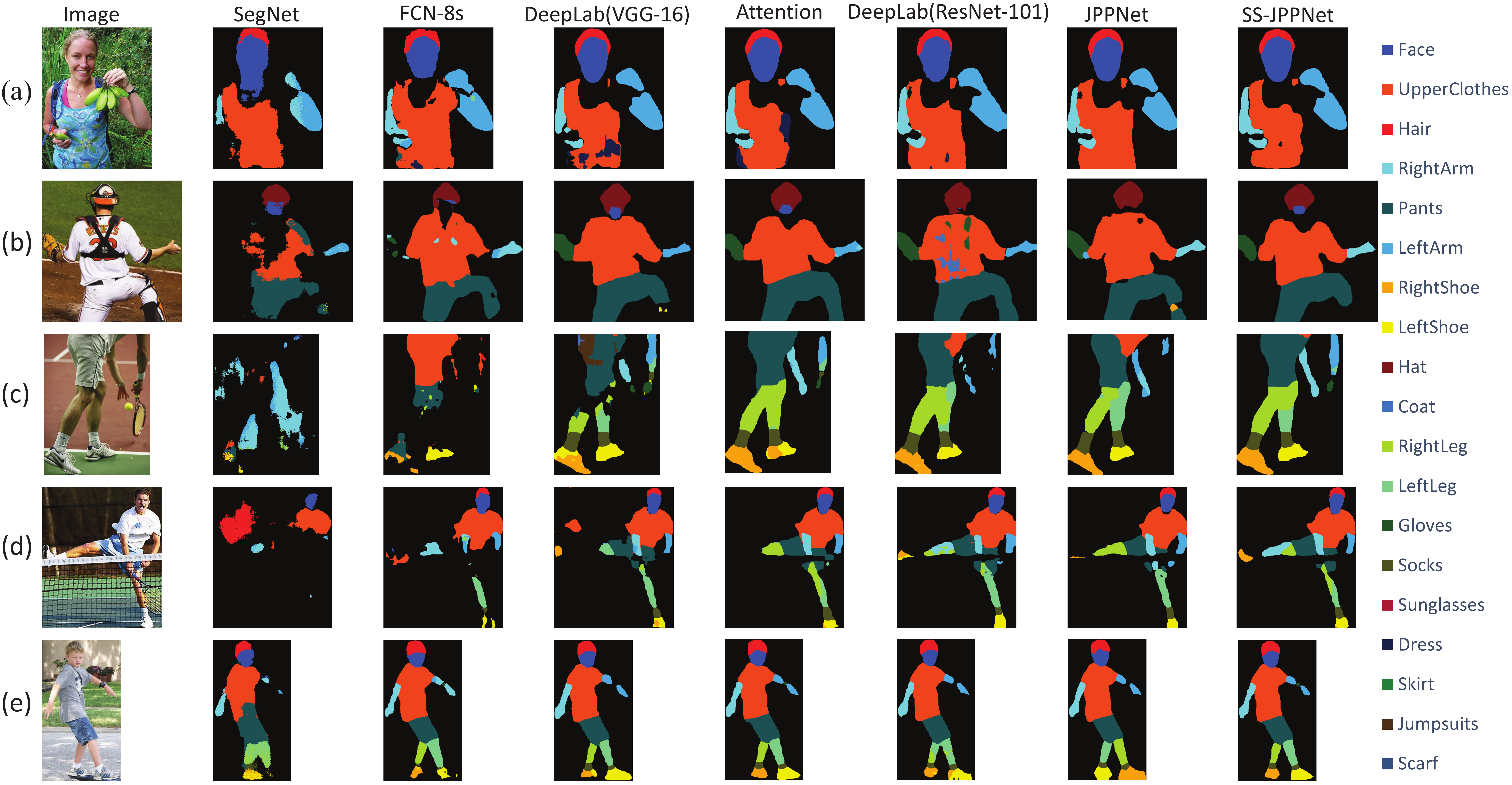}
\vspace{-4mm}
\caption{Visualized comparison of human parsing results on the LIP validation set. (a) The upper-body images. (b) The back-view images. (c) The head-missing images. (d) The images with occlusion. (e) The full-body images.}
\vspace{-2mm}
\label{fig:comparison}
\end{figure*}

\section{Empirical Study of State-of-the-art}
In this section, we analyze the performance of leading human parsing or semantic object segmentation and pose estimation approaches on our benchmark. We take advantage of our rich annotations and conduct a detailed analysis of the various factors that influence the results, such as appearance, foreshortening, and viewpoints. The goal of this analysis is to evaluate the robustness of the current approaches in various challenges for human parsing and pose estimation and identify the existing limitations to stimulate further research advances.

\subsection{Human Parsing}
In our analysis, we consider fully convolutional networks~\cite{long2014fully} (FCN-8s), a deep convolutional encoder-decoder architecture~\cite{badrinarayanan2015segnet} (SegNet), deep convolutional nets with atrous convolution and multi-scale~\cite{chen2016deeplab} (DeepLab (VGG-16), DeepLab (ResNet-101)) and an attention mechanism~\cite{chen2015attention} (Attention), which all have achieved excellent performance on semantic image segmentations in different ways and have completely available codes. For a fair comparison, we train each method on our LIP training set {until the validation performance saturates}, and we perform evaluation on the validation set and the test set. For the DeepLab methods, we remove the post-processing, dense CRFs. Following~\cite{chen2015attention,xia2015zoom}, we use the standard intersection over union (IoU) criterion and pixel-wise accuracy for evaluation.
\subsubsection{Overall Performance Evaluation}
We begin our analysis by reporting the overall human parsing performance of each approach, and the results are summarized in Table~\ref{tab: lip_val} and Table~\ref{tab: lip_test}. On the LIP validation set, among the five approaches, DeepLab (ResNet-101)~\cite{chen2016deeplab} with the deepest networks  achieves the best result of 44.80\% mean IoU. Benefitting from the attention model that softly weights the multi-scale features, Attention~\cite{chen2015attention} also performs well with 42.92\% mean IoU, whereas both FCN-8s~\cite{long2014fully} (28.29\%) and SegNet~\cite{badrinarayanan2015segnet} (18.17\%) perform significantly worse. 
Similar performance is observed on the LIP test set. The interesting outcome of this comparison is that the achieved performance is substantially lower than the current best results on other segmentation benchmarks, such as PASCAL VOC~\cite{everingham2012pascal}. This result suggests that detailed human parsing due to the small parts and diverse fine-grained labels is more challenging than object-level segmentation, which deserves more attention in the future.
\subsubsection{Performance Evaluation under Different Challenges}
We further analyze the performance of each approach with respect to the following five challenging factors: occlusion, full body, upper body, head missing and back view (see Fig.~\ref{fig: analysis_val}). We evaluate the above five approaches on the LIP validation set, which contains 4,277 images with occlusions, 5,452 full-body images, 793 upper-body images, 112 head-missing images and 661 back-view images. As expected, the performance varies when the approaches are affected by different factors. Back view is clearly the most challenging case. For example, the IoU of Attention~\cite{chen2015attention} decreases from 42.92\% to 33.50\%. The second most influential factor is the appearance of the head. The scores of all approaches are considerably lower on head-missing images than the average score on the entire set. The performance also greatly suffers from occlusions. The results of full-body images are the closest to the average level. By contrast, upper body is relatively the easiest case, where fewer semantic parts are present and the part regions are generally larger. From these results, we can conclude that the head (or face) is an important cue for the existing human parsing approaches. The probability of ambiguous results will increase if the head part disappears in the images or in the back view. Moreover, the parts or clothes on the lower body are more difficult than those on the upper body because of the existence of small labels, such as shoes or socks. In this case, the body joint structure can play an effective role in guiding human parsing.
\subsubsection{Per-class Performance Evaluation}
To discuss and analyze each of the 20 labels in the LIP dataset in more detail, we further report the performance of per-class IoU on the LIP validation set, as shown in Table~\ref{tab: val_detail}. We observe that the results with respect to labels with larger regions such as face, upper clothes, coats, and pants are considerably better than those on the small-region labels, such as sunglasses, scarf, and skirt. DeepLab (ResNet-101)~\cite{chen2016deeplab} and Attention~\cite{chen2015attention} perform better on small labels thanks to the utilization of deep networks and multi-scale features.
\subsubsection{Visualization Comparison}
The qualitative comparisons of the five approaches on our LIP validation set are visualized in Fig.~\ref{fig:comparison}. We present example parsing results of the five challenging factor scenarios. For the upper-body image (a) with slight occlusion, the five approaches perform well with few errors. For the back-view image (b), all five methods mistakenly label the right arm as the left arm. The worst results occur for the head-missing image (c). SegNet~\cite{badrinarayanan2015segnet} and FCN-8s~\cite{long2014fully} fail to recognize arms and legs, whereas DeepLab (VGG-16)~\cite{chen2016deeplab} and Attention~\cite{chen2015attention} have errors on the right and left arms, legs and shoes. Furthermore, severe occlusion (d) also greatly affects the performance. Moreover, as observed from (c) and (d), some of the results are unreasonable from the perspective of human body configuration (e.g., two shoes on one foot) because the existing approaches lack the consideration of body structures. In summary, human parsing is more difficult than general object segmentation. Particularly, human body structures should receive more attention to strengthen the ability to predict human parts and clothes with more reasonable configurations. Consequently, we consider connecting human parsing results and body joint structure to determine a better approach for human parsing.

\begin{table*}[t]
\centering
\scriptsize
\tabcolsep 0.015in 
\caption{Performance comparison in terms of per-class IoU with five state-of-the-art methods on the LIP validation set.}
\vspace{-3mm}
\label{tab: val_detail}
\begin{tabular}{cccccccccccccccccccccc}
\toprule[0.7pt]
Method                                  & hat & hair & gloves & sunglasses & u-clothes & dress & coat & socks & pants & jumpsuit & scarf & skirt & face & l-arm & r-arm & l-leg & r-leg & l-shoe & r-shoe & Bkg & Avg\\ \hline 

SegNet~\cite{badrinarayanan2015segnet}  & 26.60 & 44.01 & 0.01  & 0.00  & 34.46 & 0.00  & 15.97 & 3.59  & 33.56 & 0.01  & 0.00  & 0.00  & 52.38 & 15.30 & 24.23 & 13.82 & 13.17 & 9.26  & 6.47  & 70.62 & 18.17  \\

FCN-8s~\cite{long2014fully}             & 39.79 & 58.96 & 5.32  & 3.08  & 49.08 & 12.36 & 26.82 & 15.66 & 49.41 & 6.48  & 0.00  & 2.16  & 62.65 & 29.78 & 36.63 & 28.12 & 26.05 & 17.76 & 17.70 & 78.02 & 28.29  \\

DeepLab (VGG-16)~\cite{chen2016deeplab}        & 57.94 & 66.11 & 28.50 & 18.40 & 60.94 & 23.17 & 47.03 & 34.51 & 64.00 & 22.38 & 14.29 & 18.74 & 69.70 & 49.44 & 51.66 & 37.49 & 34.60 & 28.22 & 22.41 & 83.25 & 41.64    \\

Attention~\cite{chen2015attention}      & 58.87 & 66.78 & 23.32 & 19.48 & 63.20 & 29.63 & 49.70 & 35.23 & 66.04 & 24.73 & 12.84 & 20.41 & 70.58 & 50.17 & 54.03 & 38.35 & 37.70 & 26.20 & 27.09 & 84.00 & 42.92   \\ 

DeepLab (ResNet-101)~\cite{chen2016deeplab}      & 59.76 & 66.22 & 28.76 & \textbf{23.91} & 64.95 & \textbf{33.68} & 52.86 & 37.67 & 68.05 & 26.15 & 17.44 & \textbf{25.23} & 70.00 & 50.42 & 53.89 & 39.36 & 38.27 & 26.95 & 28.36 & 84.09 & 44.80   \\ \hline

JPPNet {(with pose info)}  & \textbf{63.55} & \textbf{70.20} & \textbf{36.16} & 23.48 & \textbf{68.15} & 31.42 & \textbf{55.65} & \textbf{44.56} & \textbf{72.19} & \textbf{28.39} & \textbf{18.76} & 25.14 & \textbf{73.36} & \textbf{61.97} & \textbf{63.88} & \textbf{58.21} & \textbf{57.99} & \textbf{44.02} & \textbf{44.09} & \textbf{86.26} & \textbf{51.37}  \\

SS-JPPNet                          & 59.75 & 67.25 & 28.95 & 21.57 & 65.30 & 29.49 & 51.92 & 38.52 & 68.02 & 24.48 & 14.92 & 24.32 & 71.01 & 52.64 & 55.79 & 40.23 & 38.80 & 28.08 & 29.03 & 84.56 & 44.73  \\

\toprule[0.7pt]
\vspace{-4mm}
\end{tabular}
\end{table*}

\begin{table*}[t]
\centering
\normalsize
\vspace{-2mm}
\caption{Comparison of human pose estimation performance with state-of-the-art methods on the LIP test set.}
\vspace{-3mm}
\label{tab: lip_test_pose}
\begin{tabular}{ccccccccc}
\toprule[0.7pt]
   Method                                       & Head  & Shoulder & Elbow & Wrist & Hip  & Knee & Ankle & Total  \\ \hline 
   ResNet-101~\cite{chen2016deeplab}            & 91.2  & 84.4     & 78.5  & 75.7  & 62.8 & 70.1 & 70.6  & 76.8   \\
   CPM~\cite{Wei_2016_CVPR}                     & 90.8  & 85.1     & 78.7  & 76.1  & 64.7 & 70.5 & 71.2  & 77.3   \\
   Hourglass~\cite{newell2016stacked}           & 91.1  & 85.3     & 78.9  & 76.2  & 65.0 & 70.2 & 72.2  & 77.6   \\
   JPPNet {(with parsing info)}  & \textbf{93.3}  & \textbf{89.3} & \textbf{84.4} & \textbf{82.5} & \textbf{70.0} & \textbf{78.3} & \textbf{77.7} & \textbf{82.7} \\
\toprule[0.7pt]
\vspace{-4mm}
\end{tabular}
\end{table*}

\begin{table*}[t]
\centering
\normalsize
\vspace{-3mm}
\caption{Comparison of human pose estimation performance with state-of-the-art methods on the LIP validation set.}
\vspace{-3mm}
\label{tab: lip_val_pose}
\begin{tabular}{ccccccccc}
\toprule[0.7pt]
   Method                                       & Head  & Shoulder & Elbow & Wrist & Hip  & Knee & Ankle & Total  \\ \hline 
   ResNet-101~\cite{chen2016deeplab}            & 91.2  & 84.3     & 78.0  & 74.9  & 62.3 & 69.5 & 71.1  & 76.5   \\
   CPM~\cite{Wei_2016_CVPR}                     & 91.1  & 85.1     & 78.7  & 75.0  & 63.7 & 69.6 & 71.7  & 77.0   \\
   Hourglass~\cite{newell2016stacked}           & 91.2  & 85.7     & 78.7  & 75.5  & 64.8 & 70.5 & 72.1  & 77.5   \\
   JPPNet {(with parsing info)} & \textbf{93.2}  & \textbf{89.3} & \textbf{84.6} & \textbf{82.2} & \textbf{69.9} & \textbf{78.0} & \textbf{77.3} & \textbf{82.5} \\
\toprule[0.7pt]
\vspace{-6mm}
\end{tabular}
\end{table*}

\begin{figure}[t]
\centering
   \includegraphics[width=1.0\linewidth]{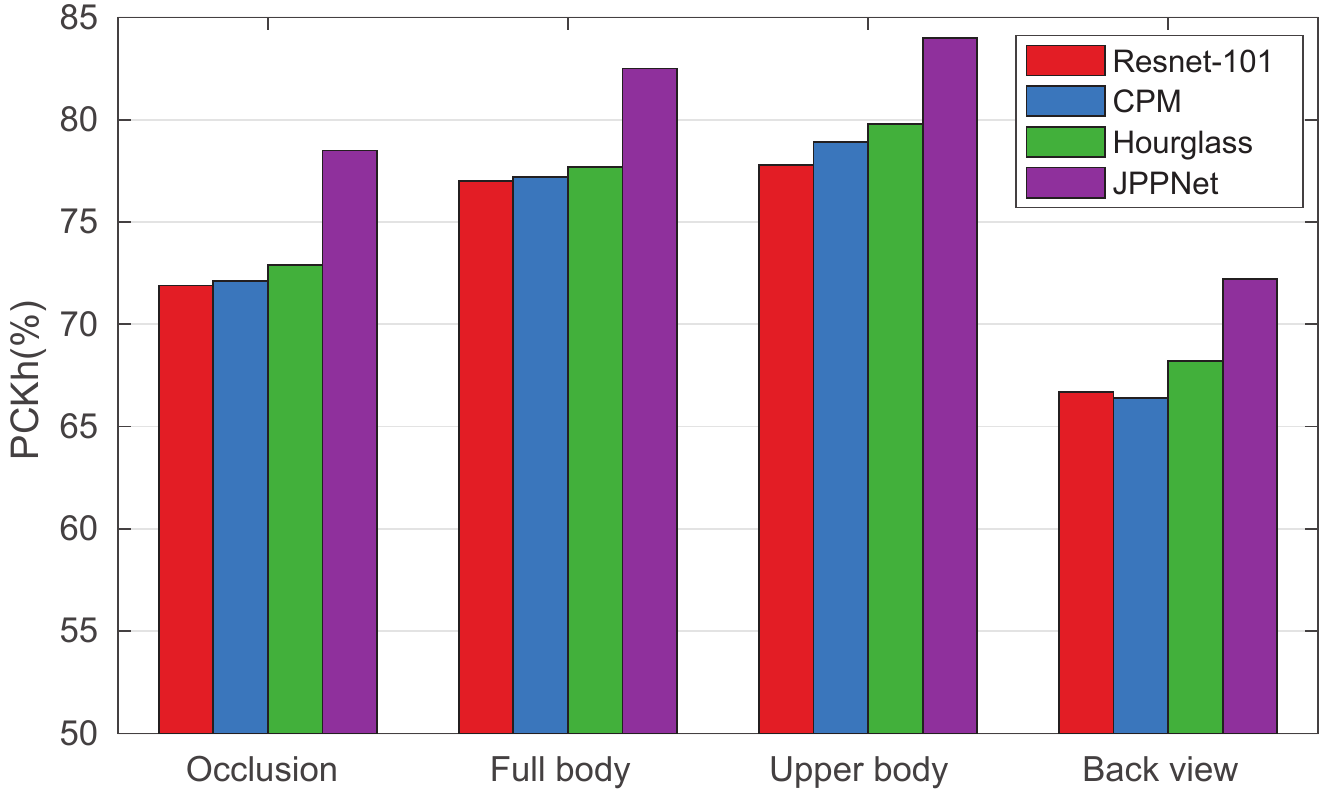}
\vspace{-8mm}
\caption{Pose estimation performance comparison evaluated on the LIP validation set with different appearances.}
\vspace{-2mm}
\label{fig: analysis_val_pose}
\end{figure}

\begin{figure*}[t]
\centering
\includegraphics[width=0.9\linewidth]{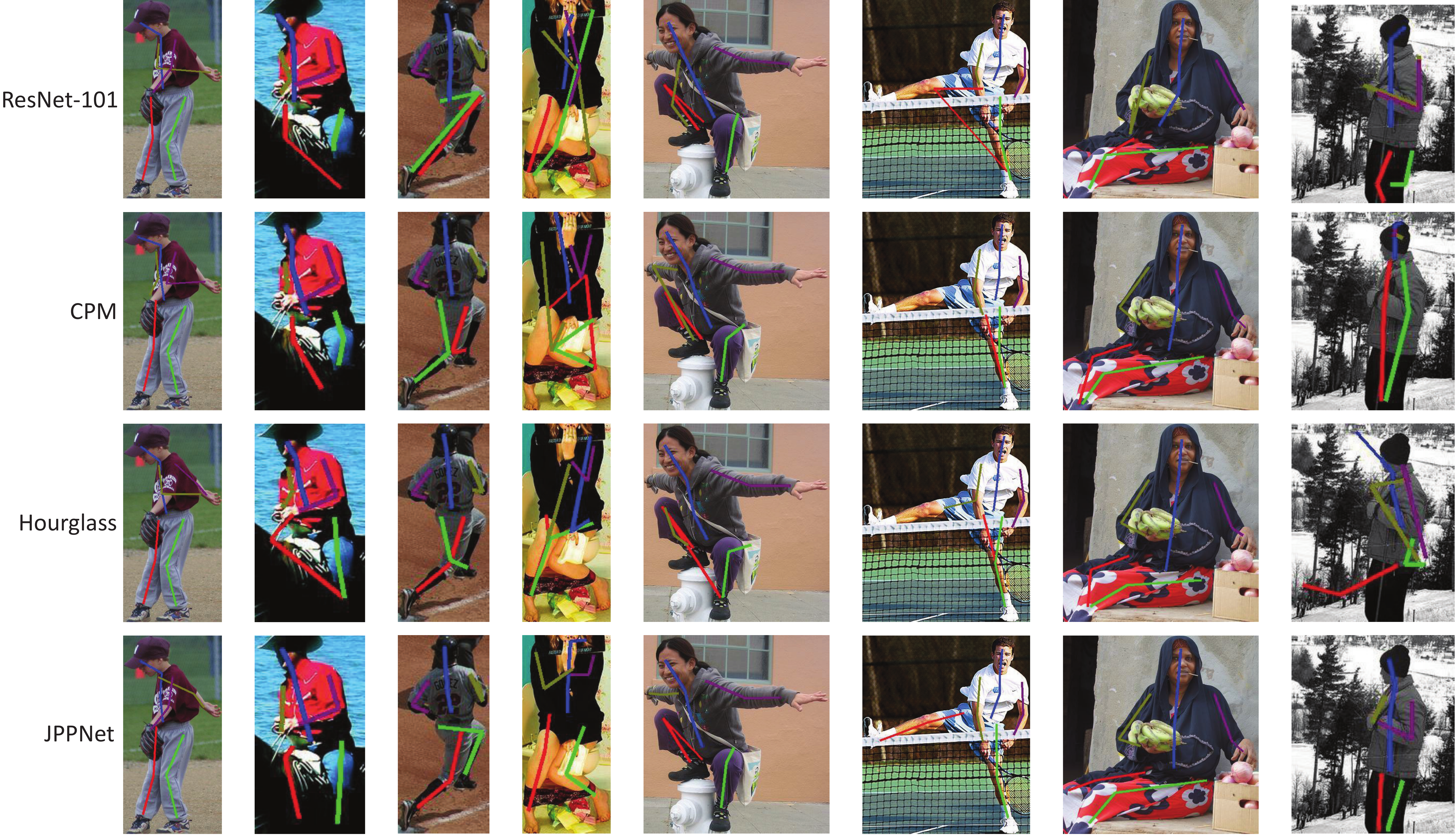}
\vspace{-4mm}
\caption{Visualized comparison of pose estimation results with three state-of-the-art methods, including ResNet-101, CPM and Hourglass on the LIP validation set.}
\vspace{-2mm}
\label{fig:comparison_pose}
\end{figure*}

\subsection{Pose Estimation}
Similarly, we consider three state-of-the-art methods for pose estimation, including a sequential convolutional architecture~\cite{Wei_2016_CVPR} (CPM) and a repeated bottom-up, top-down network~\cite{newell2016stacked} (Hourglass). ResNet-101 with atrous convolutions~\cite{chen2016deeplab} is also taken into account, for which we reserve the entire network and change the output layer to generate pose heatmaps. These approaches achieve top performance on the MPII~\cite{andriluka14cvpr} and LSP~\cite{Johnson10} datasets and can be trained on our LIP dataset using publicly available codes. Again, we train each method on our LIP training set and evaluate on the validation set and the test set. {Following MPII~\cite{andriluka14cvpr}, the evaluation metric that we used is the percentage of correct keypoints with respect to head (PCKh). PCKh considers a candidate keypoint to be localized correctly if it falls within the matching threshold which is 50\% of the head segment length. }

\subsubsection{Overall Performance Evaluation}
We again begin our analysis by reporting the overall pose estimation performance of each approach, and the results are summarized in Table~\ref{tab: lip_test_pose} and Table~\ref{tab: lip_val_pose}. On the LIP validation set, Hourglass~\cite{newell2016stacked} achieves the best result of 77.5\%
total PCKh, benefiting from their multiple hourglass modules and intermediate supervision. With a sequential composition of convolutional architectures to learn implicit spatial models, CPM~\cite{Wei_2016_CVPR} also obtains comparable performance. Interestingly, the achieved performance is substantially lower than the current best results on other pose estimation benchmarks, such as MPII~\cite{andriluka14cvpr}. This wide gap reflects the higher complexity and variability of our LIP dataset and the significant development potential of pose estimation research. Similar performance on the LIP test set is again consistent with our analysis.

\subsubsection{Performance Evaluation under Different Challenges}

We further analyze the performance of each approach with respect to the four challenging factors (see Fig.~\ref{fig: analysis_val_pose}). We leave head-missing images out because the PCKh metric depends on the head size of the person. In general and as expected, the performance decreases as the complexity increases. However, there are interesting differences. The back-view factor clearly influences the performance of all approaches the most, as the scores of all approaches decrease nearly 10\% compared to the average score on the entire set. The second most influential factor is occlusion. For example, the PCKh of Hourglass~\cite{newell2016stacked} is 4.60\% lower. These two factors are related to the visibility and orientation of heads in the images, which indicates that similar to human parsing, the existing pose estimation methods strongly depend on the contextual information of the head or face. In this case, exploring and leveraging the correlation and complementation of human parsing and pose estimation is advantageous for reducing this type of dependency. 

\subsubsection{Visualization Comparison}
The qualitative comparisons of the pose estimation results on our LIP validation set are visualized in Fig.~\ref{fig:comparison_pose}. We select some challenging images with unusual appearances, truncations and occlusions to analyze the failure cases and obtain some inspiration. First, for the persons standing or sitting sideways, the existing approaches typically failed to predict their occluded body joints, such as the right arm in Col 1, the right leg in Col 2, and the right leg in Col 7. Second, for the persons in back view or head missing,  the left and right arms (legs) of the persons are always improperly located, as with those in Cols 3 and 4. Moreover, for some images with strange appearances where some limbs of the person are very close (Cols 5 and 6), ambiguous and irrational results will be generated by these methods. In particular, the performance of the gray images (Col 8) is also far from being satisfactory. Learning from these failure cases, we believe that pose estimation should fall back on more instructional contextual information, such as the guidance from human parts with reasonable configurations. 

Summarizing the analysis of human parsing and pose estimation, it is clear that despite the strong connection of these two human-centric tasks, the intrinsic consistency between them will benefit each other. Consequently, we present a unified framework for simultaneous human parsing and pose estimation to explore this intrinsic correlation.

\begin{figure*}[t]
\centering
  \includegraphics[width=0.9\linewidth]{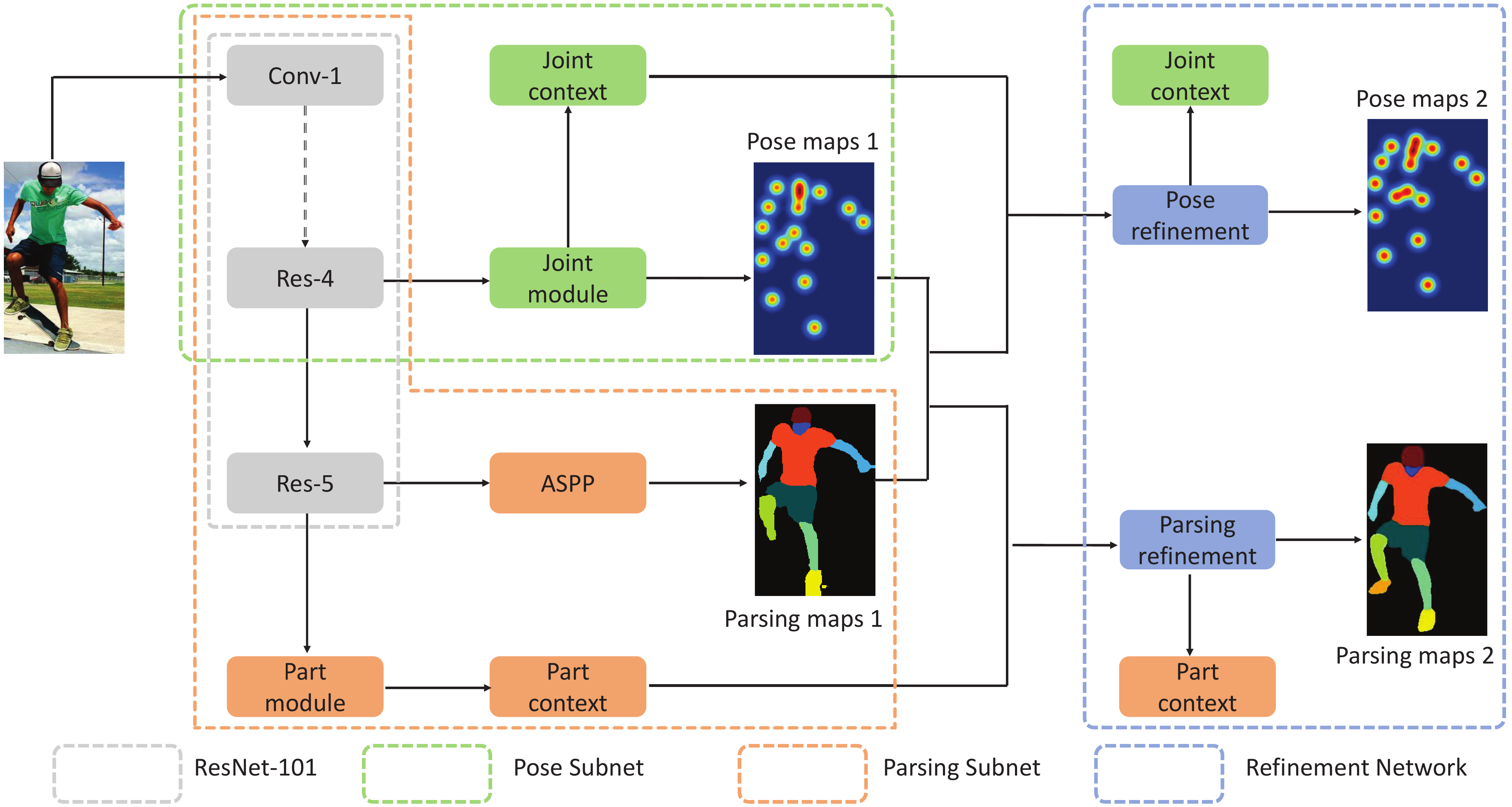}
\vspace{-4mm}
\caption{The proposed JPPNet learns to incorporate the image-level context, body joint context, body part context and refined context into a unified network, which consists of shared feature extraction, pixel-wise label prediction, keypoint heatmap prediction and iterative refinement. Given an input image, we use ResNet-101 to extract the shared feature maps. Then, a part module and a joint module are appended to capture the part context and keypoint context while simultaneously generating parsing score maps and pose heatmaps. Finally, a refinement network is performed based on the predicted maps and generated context to produce better results. To clearly observe the correlation between the parsing and pose maps, we combine pose heatmaps and parsing score maps into one map separately. For better viewing of all figures in this paper, please see the original zoomed-in color pdf file.}
\vspace{-4mm}
\label{fig:jppnet}
\end{figure*}

\section{Methods}
\subsection{Overview}
{In this section, we first summarize some insights about the limitations of existing approaches and then illustrate our joint human parsing and pose estimation network in detail.}
From the above detailed analysis, we obtain some insights into the human parsing and pose estimation tasks. 1) A major limitation of the existing human parsing approaches is the lack of consideration of human body configuration, which is mainly investigated in the human pose estimation problem. Meanwhile, the part maps produced by the detection network contain multiple contextual information and structural part relationships, which can effectively guide the regression network to predict the locations of the body joints. Human parsing and pose estimation aim to label each image with different granularities, that is, pixel-wise semantic labeling versus joint-wise structure prediction. The pixel-wise labeling can address more detailed information, whereas joint-wise structure provides more high-level structure, which means that the two tasks are complementary. 2) As learned from the existing approaches, the coarse-to-fine scheme is widely used in both parsing and pose networks to improve accuracy. For coarse-to-fine, there are two different definitions for parsing and pose tasks. For parsing or segmentation, it means using the multi-scale segmentation or attention-to-zoom scheme~\cite{chen2015attention} for more precise pixel-wise classification. Conversely, for the pose task, it indicates iterative displacement refinement, which is widely used in pose estimation~\cite{Wei_2016_CVPR}. It is reasonable to incorporate these two distinct coarse-to-fine schemes together in a unified network to further improve the parsing and pose results.

\subsection{Joint Human Parsing and Pose Estimation Network}
To utilize the coherent representation of human parsing and pose to promote each task, we propose a joint human parsing and pose estimation network, which also incorporates two distinct coarse-to-fine schemes, i.e., multi-scale features and iterative refinement, together. The framework architecture is illustrated in Fig.~\ref{fig:jppnet} and the detailed configuration is presented in Table~\ref{tab: configuration}. We denote our joint human parsing and pose estimation network as JPPNet.

In general, the basic network of the parsing framework is a deep residual network~\cite{he2015deep}, while the pose framework prefers a stacked hourglass network~\cite{newell2016stacked}. In our joint framework, we use a shared residual network to extract human image features, which is more efficient and concise. Then, we have two distinct networks to generate parsing and pose features and results. They are followed by a refinement network, which takes features and results as input to produce more accurate segmentation and joint localization.

\textbf{Feature extraction.}
We employ convolution with upsampled filters, or ``atrous convolution''~\cite{chen2016deeplab}, as a powerful tool to repurpose ResNet-101~\cite{he2015deep} in dense prediction tasks. Atrous convolution allows us to explicitly control the resolution at which feature responses are computed within DCNNs. It also effectively enlarges the field of view of filters to incorporate larger context without increasing the number of parameters or the amount of computation. The first four stages of ResNet-101 (i.e., Res-1 to Res-4) are shared in our framework. The deeper convolutional layers are different to learn for distinct tasks.

\textbf{Parsing and pose subnet.}
We use ResNet-101 with atrous convolution as the basic parsing subnet, which contains atrous spatial pyramid pooling (ASPP) as the output layer to robustly segment objects at multiple scales. ASPP probes an incoming convolutional feature layer with filters at multiple sampling rates and effective fields of view, thus capturing objects and image context at multiple scales. Furthermore, to generate the context used in the refinement stage, there are two convolutions following Res-5. For the pose subnet, we simply add several $3\times3$ convolutional layers to the fourth stage (Res-4) of ResNet-101 to generate pose features and heatmaps.

\begin{table}[t]
\centering
\scriptsize
\tabcolsep 0.06in 
\caption{The detailed configuration of our JPPNet .}
\vspace{-3mm}
\label{tab: configuration}
\begin{tabular}{ccccc}
\toprule[0.8pt]
   Component                           & Name      & Input         & Kernel size      & Channels \\
\toprule[0.8pt]
   \multirow{2}{*}{Part module}        & conv-1    & Res-5         & $3\times3$       &  512     \\
                                       & conv-2    & conv-1        & $3\times3$       &  256     \\
\toprule[0.8pt]
   \multirow{8}{*}{Joint module}       & conv-1    & Res-4         & $3\times3$       &  512     \\
                                       & conv-2    & conv-1        & $3\times3$       &  512     \\
                                       & conv-3    & conv-2        & $3\times3$       &  256     \\
                                       & conv-4    & conv-3        & $3\times3$       &  256     \\
                                       & conv-5    & conv-4        & $3\times3$       &  256     \\
                                       & conv-6    & conv-5        & $3\times3$       &  256     \\
                                       & conv-7    & conv-6        & $1\times1$       &  512     \\
                                       & conv-8    & conv-7        & $1\times1$       &  16      \\
\toprule[0.8pt]
   \multirow{9}{*}{Pose refinement}    & remap-1   & pose maps     & $1\times1$       &  128     \\
                                       & remap-2   & parsing maps  & $1\times1$       &  128     \\
                                       & \multirow{3}{*}{concat}   & remap-1          & \multirow{3}{*}{-}  & \multirow{3}{*}{512}  \\
                                       &                           & remap-2          &                     &                       \\
                                       &                           & pose context     &                     &                       \\
                                       & conv-1    & concat        & $3\times3$       &  512     \\
                                       & conv-2    & conv-1        & $5\times5$       &  256     \\
                                       & conv-3    & conv-2        & $7\times7$       &  256     \\
                                       & conv-4    & conv-3        & $9\times9$       &  256     \\
                                       & conv-5    & conv-4        & $1\times1$       &  256     \\
                                       & conv-6    & conv-5        & $1\times1$       &  16     \\ 
\toprule[0.8pt]
   \multirow{9}{*}{Parsing refinement} & remap-1   & pose maps     & $1\times1$       &  128     \\
                                       & remap-2   & parsing maps  & $1\times1$       &  128     \\
                                       & \multirow{3}{*}{concat}   & remap-1          & \multirow{3}{*}{-}  & \multirow{3}{*}{512}  \\
                                       &                           & remap-2          &                     &                       \\
                                       &                           & parsing context   &                     &                       \\
                                       & conv-1    & concat        & $3\times3$       &  512     \\
                                       & conv-2    & conv-1        & $5\times5$       &  256     \\
                                       & conv-3    & conv-2        & $7\times7$       &  256     \\
                                       & conv-4    & conv-3        & $9\times9$       &  256     \\
                                       & conv-5    & conv-4        & $1\times1$       &  256     \\
                                       & ASPP      & conv-5        & -                &  20     \\ 

\toprule[0.8pt]
\vspace{-6mm}
\end{tabular}
\end{table}

\textbf{Refinement network.}
We also design a simple but efficient refinement network, which is able to iteratively refine both parsing and pose results. We reintegrate the intermediate parsing and pose predictions back into the feature space by mapping them to a larger number of channels with an additional $1\times1$ convolution. Then, we have four convolutional layers with an incremental kernel size that varies from 3 to 9 to capture a sufficient local context and to increase the receptive field size, which is crucial for learning long-term relationships. Next is another $1\times1$ convolution to generate the features for the next refinement stage. To refine pose, we concatenate the remapped pose and parsing results and the pose features from the last stage. For parsing, we concatenate the two remapped results and parsing features and use ASPP again to generate parsing predictions. The entire joint network with refinement can be trained end-to-end, feeding the output of the former stage into the next. Following other pose estimation methods that have demonstrated strong performance with multiple iterative stages and intermediate supervision~\cite{Wei_2016_CVPR,newell2016stacked,carreira2016human}, we apply a loss upon the prediction of intermediate maps.


\begin{figure}[t]
\centering
   \includegraphics[width=1.0\linewidth]{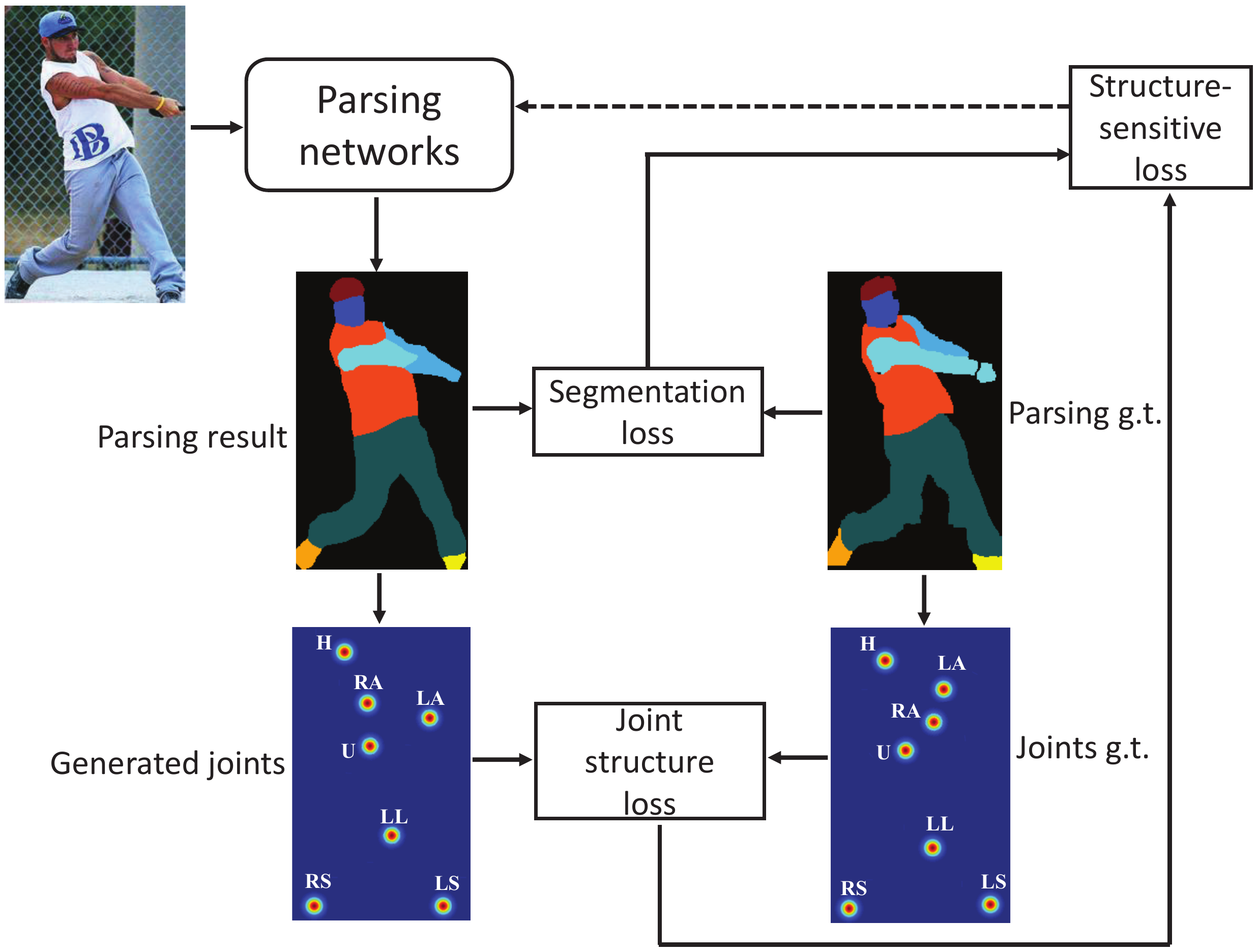}
\vspace{-7mm}
\caption{Illustration of our SS-JPPNet for human parsing. An input image goes through parsing networks, including several convolutional layers, to generate the parsing results. The generated joints and ground truths of joints represented as heatmaps are obtained by computing the center points of corresponding regions in parsing maps, including head (H), upper body (U), lower body (L), right arm (RA), left arm (LA), right leg (RL), left leg (LL), right shoe (RS), and left shoe (LS). The structure-sensitive loss is generated by weighting segmentation loss with joint structure loss. For a clear observation,  we combine nine heatmaps into one map here.}
\vspace{-2mm}
\label{fig:ss_jppnet}
\end{figure}

\subsection{Self-supervised Structure-sensitive Learning}
The joint human parsing and pose estimation network (JPPNet) leverages both pixel-wise supervision from human part annotations and high-level structural guidance from joint annotations. However, in some cases, e.g., in previous human parsing datasets, the joint annotations may not be available. In this section, we show that high-level human structure cues can still help the human parsing task even without explicit supervision from manual annotations. We simplify our JPPNet and propose a novel self-supervised structure-sensitive learning for human parsing, which introduces a self-supervised structure-sensitive loss to evaluate the quality of the predicted parsing results from a joint structure perspective, as illustrated in Fig.~\ref{fig:ss_jppnet}.

Specifically, in addition to using the traditional pixel-wise annotations as the supervision, we generate the approximated human joints directly from the parsing annotations, which can also guide human parsing training. For the purpose of explicitly enforcing the produced parsing results to be semantically consistent with the human joint structures, we treat the joint structure loss as a weight of segmentation loss, which becomes our structure-sensitive loss.

\begin{figure}[t]
\centering
\includegraphics[width=0.9\linewidth]{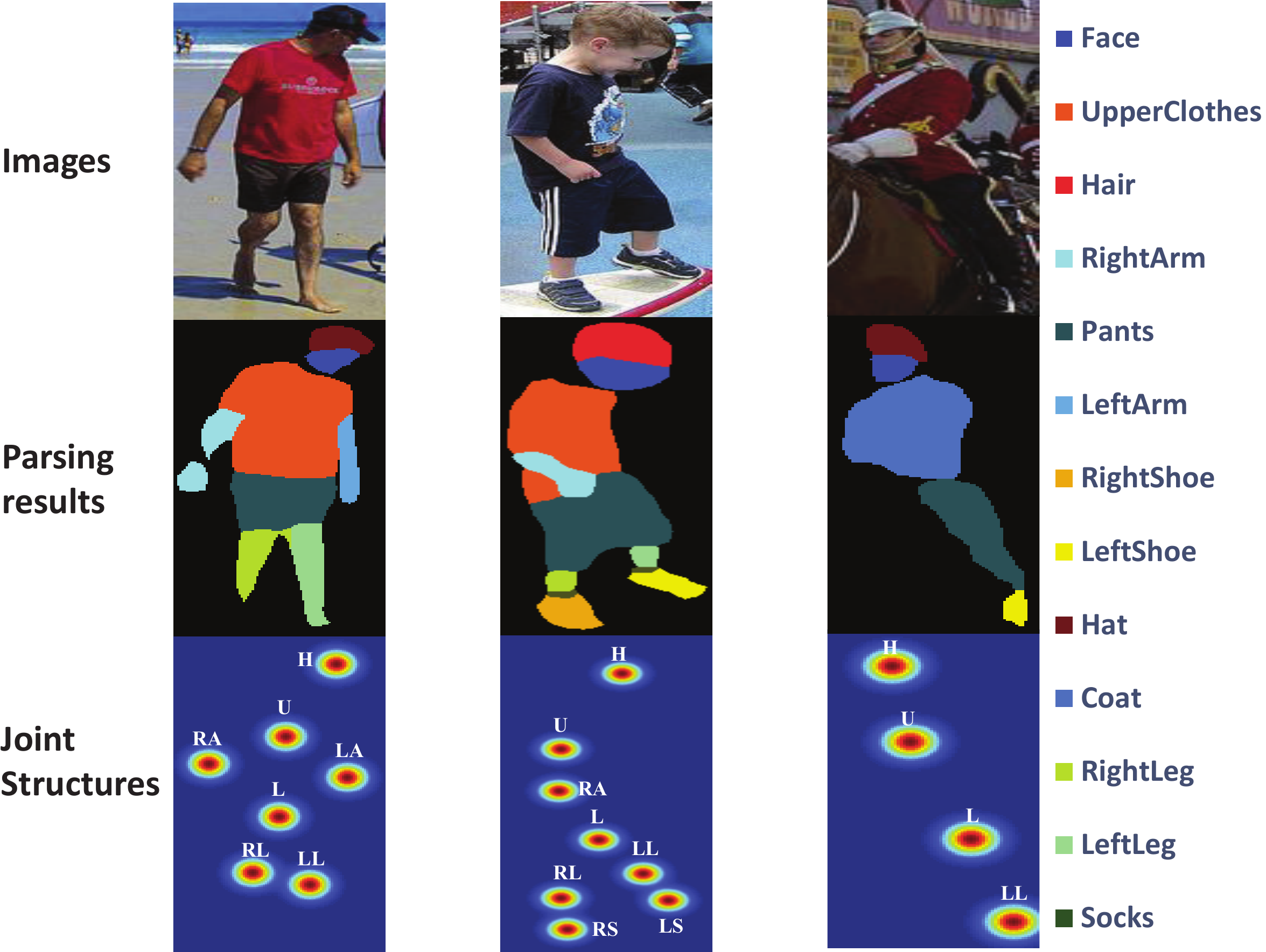}
\vspace{-4mm}
\caption{Some examples of self-supervised human joints generated from our parsing results for different bodies.}
\vspace{-2mm}
\label{fig:heatmap_example}
\end{figure}

\textbf{Self-supervised Structure-sensitive Loss: }
Generally, for the human parsing task, no other extensive information is provided except the pixel-wise annotations. This situation means that rather than using augmentative information, we have to find a structure-sensitive supervision from the parsing annotations. Because the human parsing results are semantic parts with pixel-level labels, we attempt to explore pose information contained in human parsing results. We define 9 joints to construct a pose structure, which are the centers of the regions of head, upper body, lower body, left arm, right arm, left leg, right leg, left shoe and right shoe. The head regions are generated by merging the parsing labels of hat, hair, sunglasses and face. Similarly, upper clothes, coat and scarf are merged to be upper body, and pants and skirt are merged for lower body. The remaining regions can also be obtained by the corresponding labels. Some examples of generated human joints for different humans are shown in Fig.~\ref{fig:heatmap_example}. Following~\cite{Pfister15a}, for each parsing result and corresponding ground truth, we compute the center points of regions to obtain joints represented as heatmaps for training more smoothly. Then, we use the Euclidean metric to evaluate the quality of the generated joint structures, which also reflects the structure consistency between the predicted parsing results and the ground truth. Finally, the pixel-wise segmentation loss is weighted by the joint structure loss, which becomes our structure-sensitive loss. Consequently, the overall human parsing networks become self-supervised with the structure-sensitive loss. 

Formally, given an image $I$, we define a list of joint configurations $C^{P}_I = \{c^{p}_i|i\in[1,N]\}$, where $c^{p}_i$ is the heatmap of the i-th joint computed according to the parsing result map. Similarly, $C^{GT}_I = \{c^{gt}_i|i\in[1,N]\}$, which is obtained from the corresponding parsing ground truth. Here, $N$ is a variable decided by the human bodies in the input images, and it is equal to 9 for a full-body image. For the joints missed in the image, we simply replace the heatmaps with maps filled with zeros. The joint structure loss is the Euclidean (L2) loss, which is calculated as follows:
\begin{equation}
L_{\text{Joint}} = \frac{1}{2N} \sum \limits_{i=1}^N \| c^{p}_i - c_i^{gt} \|_2^2
\end{equation}
The final structure-sensitive loss, denoted as $L_{\text{structure}}$, is the combination of the joint structure loss and the parsing segmentation loss, and it is calculated as follows:
\begin{equation}
L_{\text{Structure}} = L_{\text{Joint}} \cdot L_{\text{Parsing}}
\end{equation}
where $L_{\text{Parsing}}$ is the pixel-wise softmax loss calculated based on the parsing annotations.

We name this learning strategy ``self-supervised'' because the above structure-sensitive loss can be generated from existing parsing results without any extra information. Our self-supervised structure-sensitive JPPNet (SS-JPPNet) thus has excellent adaptability and extensibility, which can be injected into any advanced network to help incorporate rich high-level knowledge about human joints from a global perspective.

\section{Experiments}
\subsection{Experimental Settings}
\textbf{Network architecture: }
We utilize the publicly available model DeepLab (ResNet-101)~\cite{chen2016deeplab} as the basic architecture of our JPPNet, which employs atrous convolution, multi-scale inputs with max-pooling to merge the results from all scales, and atrous spatial pyramid pooling. For SS-JPPNet, our basic network is Attention~\cite{chen2015attention} due to its leading accuracy and competitive efficiency.

\textbf{Training: }
To train JPPNet, the input image is scaled to $384 \times 384$. We first train ResNet-101 on the human parsing task for 30 epochs using the pre-trained models and networks settings from~\cite{chen2016deeplab}. Then, we train the joint framework end-to-end for another 30 epochs. We apply data augmentation, including randomly scaling the input images (from 0.75 to 1.25), randomly cropping and randomly left-right flipping during training.

When training SS-JPPNet, we use the pre-trained models and network settings provided by DeepLab~\cite{chen2016deeplab}. The scale of the input images is fixed as $321 \times 321$ for training networks based on Attention~\cite{chen2015attention}. Two training steps are employed to train the networks. First, we train the basic network on our LIP dataset for 30 epochs. Then, we perform the ``self-supervised" strategy to fine-tune our model with the structure-sensitive loss. We fine-tune the networks for approximately 20 epochs. We use both human parsing and pose annotations to train JPPNet and only parsing labels for SS-JPPNet.

\textbf{Inference: }
To stabilize the predictions, we perform inference on multi-scale inputs (with scales = {0.75, 0.5, 1.25}) and also left-right flipped images. In particular, we compute as the final result the average probabilities from each scale and flipped images, which is the same for predicting both parsing and pose. The difference is that we utilize predictions of all stages for parsing, but for pose, we only use the results of the last stage.

\begin{table}[t]
\centering
\scriptsize
\caption{Human parsing performance comparison in terms of mean IoU. Left: different test sets. Right: different sizes of objects.}
\vspace{-3mm}
\label{tab: lip_size}
\begin{tabular}{c|cc|ccc}
\toprule[0.5pt]
Method                                     & ATR     & LIP      & small   & medium   & large     \\ \hline 
SegNet~\cite{badrinarayanan2015segnet}     & 15.79   & 21.79    & 16.53   & 18.58    & 18.18     \\
FCN-8s~\cite{long2014fully}                & 34.44   & 32.28    & 22.37   & 29.41    & 28.09     \\ 
DeepLab (VGG-16)~\cite{chen2016deeplab}     & 48.64   & 43.97    & 28.77   & 40.74    & 43.02    \\
Attention~\cite{chen2015attention}         & 49.35   & 45.38    & 31.71   & 41.61    & 44.90     \\ 
DeepLab (ResNet-101)~\cite{chen2016deeplab} & 53.28   & 46.99    & 31.70   & 43.14    & 47.62    \\ \hline
JPPNet  (with pose info)   & \textbf{54.45} & \textbf{53.99} & \textbf{44.56} & \textbf{50.52} & \textbf{52.58}    \\
SS-JPPNet                            & 52.69   & 46.85    & 33.48   & 43.12    & 46.73     \\
\toprule[0.5pt]
\end{tabular}
\end{table}

\begin{table}[t]
\centering
\scriptsize
\tabcolsep 0.025in 
\vspace{-2mm}
\caption{Comparison of human parsing performance with four state-of-the-art methods on the PASCAL-Person-Part dataset~\cite{chen2014detect}.}
\vspace{-3mm}
\label{tab: pascal}
\begin{tabular}{ccccccccc}
\toprule[0.7pt]
   Method                                     &  head  &  torso  &  u-arms &  l-arms &  u-legs &  l-legs &  Bkg   &  Avg    \\ \hline
   DeepLab-LargeFOV~\cite{chen2016deeplab}      & 78.09  &  54.02  &  37.29  &  36.85  &  33.73  &  29.61  &  92.85 &  51.78  \\
   HAZN~\cite{xia2015zoom}                       & 80.79  &  59.11  &  43.05  &  42.76  &  38.99  &  34.46  &  93.59 &  56.11  \\  
   Attention~\cite{chen2015attention}   & 81.47  &  59.06  &  44.15  &  42.50  &  38.28  &  35.62  &  93.65 &  56.39  \\ 
   LG-LSTM~\cite{liang2015semantic}              & 82.72  &  60.99  &  45.40  &  \textbf{47.76}  &  \textbf{42.33}  &  37.96  &  88.63 &  57.97  \\ \hline
   SS-JPPNet                      & \textbf{83.26}  &  \textbf{62.40}  &  \textbf{47.80}  &  45.58  &  42.32 &  \textbf{39.48}  &  \textbf{94.68} 
                                        &  \textbf{59.36}   \\
\toprule[0.7pt]
\vspace{-6mm}
\end{tabular}
\end{table}

\begin{table*}[t]
\centering
\normalsize
\vspace{-2mm}
\caption{Comparison of human pose estimation performance of the models trained on the LIP training set and evaluated on the MPII training set (11431 single person images).}
\vspace{-3mm}
\label{tab: pose_mpii}
\begin{tabular}{ccccccccc}
\toprule[0.7pt]
   Method                                       & Head  & Shoulder & Elbow & Wrist & Hip  & Knee & Ankle & Total  \\ \hline 
   ResNet-101~\cite{chen2016deeplab}            & 89.2  & 86.7     & 79.5  & 77.7  & 75.5 & 66.7 & 61.8  & 77.9   \\
   CPM~\cite{Wei_2016_CVPR}                     & 86.6  & 83.6     & 75.8  & 72.1  & 70.9 & 62.0 & 59.1  & 74.0   \\
   Hourglass~\cite{newell2016stacked}           & 86.4  & 84.7     & 77.5  & 73.9  & 74.0 & 63.3 & 58.4  & 75.2   \\
   JPPNet (with parsing info)  & \textbf{90.4}  & \textbf{91.7} & \textbf{86.4} & \textbf{84.0} & \textbf{82.5} & \textbf{76.5} & \textbf{71.3} & \textbf{84.1} \\
\toprule[0.7pt]
\vspace{-6mm}
\end{tabular}
\end{table*}

\begin{table*}[t]
\centering
\normalsize
\vspace{-2mm}
\caption{Human pose estimation comparison between different variants of the proposed JPPNet on the LIP test set using the PCKh metric.}
\vspace{-3mm}
\label{tab: ablation_pose}
\begin{tabular}{ccccccccc}
\toprule[0.7pt]
   Method             & Head          & Shoulder & Elbow & Wrist & Hip   & Knee & Ankle & Total  \\ \hline 
   Joint              & 93.1          & 87.2     & 81.1  & 79.3  & 66.8  & 72.6 & 72.9  & 79.6   \\
   Joint + MSC        & 92.9          & 88.1     & 82.8  & 81.0  & 67.8  & 74.3 & 75.7  & 80.9   \\
   Joint + S1         & \textbf{93.5} & 88.8     & 83.6  & 81.6  & 70.2  & 76.4 & 76.8  & 82.1   \\
   Joint + MSC + S1   & 93.4 & 88.9   & 84.2     & 82.1  & \textbf{70.8} & 77.2 & 77.3  & 82.5   \\
   Joint + MSC + S2   & 93.3          & \textbf{89.3} & \textbf{84.4} & \textbf{82.5} & 70.0 & \textbf{78.3} & \textbf{77.7} & \textbf{82.7} \\
\toprule[0.7pt]
\vspace{-6mm}
\end{tabular}
\end{table*}

\subsection{Results and Comparisons}

\subsubsection{Human Parsing}
We compare our proposed approach with the strong baselines on the LIP dataset, and we further evaluate SS-JPPNet on another public human parsing dataset.

\textbf{LIP dataset:}
We report the results and the comparisons with five state-of-the-art methods on the LIP validation set and test set in Table~\ref{tab: lip_val} and Table~\ref{tab: lip_test}. On the validation set, the proposed JPPNet framework improves the best performance from 44.80\% to 51.37\%. The simplified architecture can also provide a substantial enhancement in average IoU: 3.09\% better than DeepLab (VGG-16)~\cite{chen2016deeplab} and 1.81\% better than Attention~\cite{chen2015attention}. On the test set, the JPPNet also considerably outperforms the other baselines. This superior performance achieved by our methods demonstrates the effectiveness of our joint parsing and pose networks, which incorporate the body joint structure into the pixel-wise prediction.

In Fig.~\ref{fig: analysis_val}, we show the results with respect to the different challenging factors on our LIP validation set. With our unified framework that models the contextual information of body parts and joints, the performance of all kinds of types is improved, which demonstrates that human joint structure is conducive for the human parsing task.

We further report per-class IoU on the LIP validation set to verify the detailed effectiveness of our approach, as presented in Table~\ref{tab: val_detail}. With the consideration of human body joints, we achieved the best performance on almost all the classes. As observed from the reported results, the proposed JPPNet significantly improves the performance of the labels such as arms, legs, and shoes, which demonstrates its ability to refine the ambiguity of left and right. Furthermore, the labels covering small regions such as socks, and gloves are better predicted with higher IoUs. This improvement also demonstrates the effectiveness of the unified framework, particularly for small labels.

For a better understanding of our LIP dataset, {we train all methods on LIP and evaluate them on ATR~\cite{Co-CNN}, as reported in Table~\ref{tab: lip_size} (left). As ATR contains 18 categories while LIP has 20, we test the models on the 16 common categories (hat, hair, sunglasses, upper clothes, dress, pants, scarf, skirt, face, right arm, left arm, right leg, left leg, right shoe, left shoe, and background). } In general, the performance on ATR is better than those on LIP because the LIP dataset contains instances with more diverse poses, appearance patterns, occlusions and resolution issues, which is more consistent with real-world situations. 

Following the MSCOCO dataset~\cite{DBLP:journals/corr/LinMBHPRDZ14}, we have conducted an empirical analysis on different object sizes, i.e., small ($area < 153^2$), medium ($153^2 \leq area < 321^2$) and large ($area \geq 321^2$). The results of the five baselines and the proposed methods are reported in Table~\ref{tab: lip_size} (right). As shown, our methods show substantially superior performance for different sizes of objects, thus further demonstrating the advantage of incorporating the human body structure into the parsing model.

\textbf{PASCAL-Person-Part dataset~\cite{chen2014detect}.}
The public PASCAL-Person-Part dataset with 1,716 images for training and 1,817 for testing focuses on the human part segmentation annotated by~\cite{chen2014detect}. Following~\cite{chen2015attention,xia2015zoom}, the annotations are merged to be six person part classes and one background class, which are head, torso, upper / lower arms and upper / lower legs. {We train and evaluate all methods using the training and testing data in PASCAL-Person-Part dataset~\cite{chen2014detect}.} Table~\ref{tab: pascal} shows the performance of our model and comparisons with four state-of-the-art methods on the standard IoU criterion. Our SS-JPPNet can significantly outperform the four baselines. For example, our best model achieves 59.36\% IoU, which is 7.58\% better than DeepLab-LargeFOV~\cite{chen2016deeplab} and 2.97\% better than Attention~\cite{chen2015attention}. 
This large improvement demonstrates that our self-supervised strategy is significantly beneficial for the human parsing task.

\subsubsection{Pose Estimation}
\textbf{LIP dataset:}
Table~\ref{tab: lip_test_pose} and Table~\ref{tab: lip_val_pose} report the comparison of the PCKh performance of our JPPNet and previous state-of-the-art at a normalized distance of 0.5. On the LIP test set, our method achieves state-of-the-art PCKh scores of 82.7\%. In particular, for the most challenging body parts, e.g., hip and ankle, our method achieves 5.0\% and 5.5\% improvements compared with the closest competitor, respectively. Similar improvements also occur on the validation set.

We present the results with respect to different challenging factors on our LIP validation set in Fig.~\ref{fig: analysis_val_pose}. As expected, with our unified architecture, the results of all different appearances become better, thus demonstrating the positive effects of the human parsing to pose estimation.

\textbf{MPII Human Pose dataset~\cite{andriluka14cvpr}:}
To be more convincing, we also perform evaluations on the MPII dataset. The MPII dataset contains approximately 25,000 images, where each person is annotated with 16 joints. The images are extracted from YouTube videos, where the contents are everyday human activities. There are 18079 images in the training set, including 11431 single person images. We evaluate the models trained on our LIP training set and test on these 11431 single person images from MPII, as presented in Table~\ref{tab: pose_mpii}. The distance between our approach and others provides evidence of the higher generalization ability of our proposed JPPNet model.

\subsection{Ablation Studies of JPPNet}
We further evaluate the effectiveness of our two coarse-to-fine schemes of JPPNet, including the multi-scale features and iterative refinement. {``Joint" denotes the JPPNet without multi-scale features (``MSC'') or refinement networks. ``S1'' means one stage refinement and ``S2'' is noted for two stages.} The human parsing and pose estimation results are shown in Table~\ref{tab: ablation_parsing} and Table~\ref{tab: ablation_pose}. From the comparisons, we can learn that multi-scale features greatly improve for human parsing but slightly for pose estimation. However, pose estimation considerably benefits from iterative refinement, which is not quite helpful for human parsing, as two stage refinements will decrease the parsing performance.

\begin{table}[t]
\centering
\scriptsize
\caption{Human parsing comparison between different variants of the proposed JPPNet on the LIP test set.}
\vspace{-3mm}
\label{tab: ablation_parsing}
\begin{tabular}{cccc}
\toprule[0.7pt]
   Method               & Overall accuracy & Mean accuracy   & Mean IoU  \\ \hline 
   Joint                & 86.10            & 59.64           & 49.48     \\
   Joint + MSC          & 86.18            & 61.40           & 50.83     \\
   Joint + S1           & 86.09            & 57.95           & 49.58     \\ 
   Joint + MSC + S1     & \textbf{86.48}   & \textbf{62.25}  & \textbf{51.36}     \\ 
   Joint + MSC + S2     & 86.42            & 61.12           & 50.64     \\
\toprule[0.7pt]
\vspace{-2mm}
\end{tabular}
\end{table}

\subsection{Qualitative Comparison}
\textbf{Human parsing:}
The qualitative comparisons of the parsing results on the LIP validation set are visualized in Fig.~\ref{fig:comparison}. As can be observed from these visual comparisons, our methods output more semantically meaningful and precise predictions than the other five methods despite the existence of large appearance and position variations. Taking (b) and (c) for example, our approaches can also successfully handle the confusing labels, such as left arm versus right arm and left leg versus right leg. These regions with similar appearances can be recognized and separated by the guidance from joint structure information. For the most difficult head-missing image (c), the left shoe, right shoe and legs are excellently corrected by our JPPNet approach. In general, by effectively exploiting human body joint structure, our approaches output more reasonable results for confusing labels on the human parsing task.

\textbf{Pose estimation:}
The qualitative comparisons of pose results on the LIP validation set are presented in Fig.~\ref{fig:comparison_pose}. In Section 4.2.3, we summarize some challenging cases that cause considerable trouble for the previous pose estimation approaches. In contrast, by jointly modeling human parsing and pose estimation, our model can effectively avoid the cumbersome obstacles such sideways, occlusion or other erratic postures, thus leading to more promising and reasonably remarkable results.

Finally, we want to emphasize that our goal is to explore the intrinsic correlation between human parsing and pose estimation. For this purpose, we propose JPPNet, which is a unified model built upon two distinct coarse-to-fine schemes. Separating our framework into different components leads to inferior results, as demonstrated in Table~\ref{tab: ablation_parsing} and Table~\ref{tab: ablation_pose}. Although we use more annotations than methods for individual tasks, the promising results of our framework verify that human parsing and pose estimation are essentially complementary; thus, performing the two tasks simultaneously will enhance the performance of each task.

\section{Conclusion}
In this work, we presented ``Look into Person (LIP)", a large-scale human parsing and pose estimation dataset and a carefully designed benchmark to spark progress in human-centric tasks. LIP contains 50,462 images, which are richly labeled with 19 semantic part labels and 16 body joints. It surpasses existing human parsing and pose estimation datasets in terms of scale and richness of annotations. Moreover, we proposed a joint human parsing and pose estimation network to explore the intrinsic connection of the two tasks. The extensive results clearly demonstrate the effectiveness of the proposed approaches. The datasets, code and models are available at \url{http://www.sysu-hcp.net/lip/}.



\section*{Acknowledgements}

This work was supported by State Key Development Program under Grant 2016YFB1001004, the National Natural Science Foundation of China under Grant 61622214 and Grant U1611461, the Guangdong Natural Science Foundation Project for Research Teams under Grant 2017A030312006, and the Guangdong Science and Technology Planning
Program under Grant 2017B010116001.

\ifCLASSOPTIONcaptionsoff
  \newpage
\fi

{\small
\bibliographystyle{ieee}
\bibliography{egbib}

\begin{thebibliography}{10}\itemsep=-1pt

\bibitem{andriluka14cvpr}
M.~Andriluka, L.~Pishchulin, P.~Gehler, and B.~Schiele.
\newblock 2d human pose estimation: New benchmark and state of the art
  analysis.
\newblock In {\em CVPR}, June 2014.

\bibitem{andriluka2009pictorial}
M.~Andriluka, S.~Roth, and B.~Schiele.
\newblock Pictorial structures revisited: People detection and articulated pose
  estimation.
\newblock In {\em Computer Vision and Pattern Recognition, 2009. CVPR 2009.
  IEEE Conference on}, pages 1014--1021. IEEE, 2009.

\bibitem{badrinarayanan2015segnet}
V.~Badrinarayanan, A.~Kendall, and R.~Cipolla.
\newblock Segnet: A deep convolutional encoder-decoder architecture for image
  segmentation.
\newblock In {\em CVPR}, 2015.

\bibitem{bulat2016human}
A.~Bulat and G.~Tzimiropoulos.
\newblock Human pose estimation via convolutional part heatmap regression.
\newblock In {\em ECCV}, 2016.

\bibitem{carreira2016human}
J.~Carreira, P.~Agrawal, K.~Fragkiadaki, and J.~Malik.
\newblock Human pose estimation with iterative error feedback.
\newblock In {\em Proceedings of the IEEE Conference on Computer Vision and
  Pattern Recognition}, pages 4733--4742, 2016.

\bibitem{chen2014semantic}
L.-C. Chen, G.~Papandreou, I.~Kokkinos, K.~Murphy, and A.~L. Yuille.
\newblock {Semantic Image Segmentation with Deep Convolutional Nets and Fully
  Connected CRFs}.
\newblock {\em In ICLR}, 2015.

\bibitem{chen2016deeplab}
L.-C. Chen, G.~Papandreou, I.~Kokkinos, K.~Murphy, and A.~L. Yuille.
\newblock Deeplab: Semantic image segmentation with deep convolutional nets,
  atrous convolution, and fully connected crfs.
\newblock {\em arXiv preprint arXiv:1606.00915}, 2016.

\bibitem{chen2015attention}
L.-C. Chen, Y.~Yang, J.~Wang, W.~Xu, and A.~L. Yuille.
\newblock Attention to scale: Scale-aware semantic image segmentation.
\newblock In {\em CVPR}, 2016.

\bibitem{chen2014detect}
X.~Chen, R.~Mottaghi, X.~Liu, S.~Fidler, R.~Urtasun, et~al.
\newblock Detect what you can: Detecting and representing objects using
  holistic models and body parts.
\newblock In {\em CVPR}, 2014.

\bibitem{Chen_NIPS14}
X.~Chen and A.~Yuille.
\newblock Articulated pose estimation by a graphical model with image dependent
  pairwise relations.
\newblock In {\em NIPS}, 2014.

\bibitem{Chu_2017_CVPR}
X.~Chu, W.~Yang, W.~Ouyang, C.~Ma, A.~L. Yuille, and X.~Wang.
\newblock Multi-context attention for human pose estimation.
\newblock In {\em The IEEE Conference on Computer Vision and Pattern
  Recognition (CVPR)}, July 2017.

\bibitem{dantone2013human}
M.~Dantone, J.~Gall, C.~Leistner, and L.~Van~Gool.
\newblock Human pose estimation using body parts dependent joint regressors.
\newblock In {\em Computer Vision and Pattern Recognition}, pages 3041--3048,
  2013.

\bibitem{dong2014towards}
J.~Dong, Q.~Chen, X.~Shen, J.~Yang, and S.~Yan.
\newblock Towards unified human parsing and pose estimation.
\newblock In {\em CVPR}, 2014.

\bibitem{Dongparsing13}
J.~Dong, Q.~Chen, W.~Xia, Z.~Huang, and S.~Yan.
\newblock A deformable mixture parsing model with parselets.
\newblock In {\em ICCV}, 2013.

\bibitem{everingham2012pascal}
M.~Everingham, L.~Van~Gool, C.~K. Williams, J.~Winn, and A.~Zisserman.
\newblock The pascal visual object classes challenge 2010 (voc2010) results,
  2010.

\bibitem{gan2016concepts}
C.~Gan, M.~Lin, Y.~Yang, G.~de~Melo, and A.~G. Hauptmann.
\newblock Concepts not alone: Exploring pairwise relationships for zero-shot
  video activity recognition.
\newblock In {\em AAAI}, 2016.

\bibitem{he2015deep}
K.~He, X.~Zhang, S.~Ren, and J.~Sun.
\newblock Deep residual learning for image recognition.
\newblock In {\em CVPR}, 2016.

\bibitem{Jhuang:ICCV:2013}
H.~Jhuang, J.~Gall, S.~Zuffi, C.~Schmid, and M.~J. Black.
\newblock Towards understanding action recognition.
\newblock In {\em International Conf. on Computer Vision (ICCV)}, pages
  3192--3199, Dec. 2013.

\bibitem{Johnson10}
S.~Johnson and M.~Everingham.
\newblock Clustered pose and nonlinear appearance models for human pose
  estimation.
\newblock In {\em Proceedings of the British Machine Vision Conference}, 2010.
\newblock doi:10.5244/C.24.12.

\bibitem{johnson2011learning}
S.~Johnson and M.~Everingham.
\newblock Learning effective human pose estimation from inaccurate annotation.
\newblock In {\em Computer Vision and Pattern Recognition (CVPR), 2011 IEEE
  Conference on}, pages 1465--1472. IEEE, 2011.

\bibitem{kalantidis2013getting}
Y.~Kalantidis, L.~Kennedy, and L.-J. Li.
\newblock Getting the look: clothing recognition and segmentation for automatic
  product suggestions in everyday photos.
\newblock In {\em ACM conference on International conference on multimedia
  retrieval}, pages 105--112, 2013.

\bibitem{liang2015proposal}
X.~Liang, L.~Lin, Y.~Wei, X.~Shen, J.~Yang, and S.~Yan.
\newblock Proposal-free network for instance-level object segmentation.
\newblock {\em IEEE Trans. on Pattern Analysis and Machine Intelligence}, 2018.

\bibitem{ATR}
X.~Liang, S.~Liu, X.~Shen, J.~Yang, L.~Liu, J.~Dong, L.~Lin, and S.~Yan.
\newblock Deep human parsing with active template regression.
\newblock {\em TPAMI}, 2015.

\bibitem{liang2015towards}
X.~Liang, S.~Liu, Y.~Wei, L.~Liu, L.~Lin, and S.~Yan.
\newblock Towards computational baby learning: A weakly-supervised approach for
  object detection.
\newblock In {\em ICCV}, 2015.

\bibitem{liang2015semantic}
X.~Liang, X.~Shen, D.~Xiang, J.~Feng, L.~Lin, and S.~Yan.
\newblock Semantic object parsing with local-global long short-term memory.
\newblock In {\em CVPR}, 2016.

\bibitem{Co-CNN}
X.~Liang, C.~Xu, X.~Shen, J.~Yang, S.~Liu, J.~Tang, L.~Lin, and S.~Yan.
\newblock Human parsing with contextualized convolutional neural network.
\newblock In {\em ICCV}, 2015.

\bibitem{LinASPL}
L.~Lin, K.~Wang, D.~Meng, W.~Zuo, and L.~Zhang.
\newblock Active self-paced learning for cost-effective and progressive face
  identification.
\newblock {\em IEEE Trans. on Pattern Analysis and Machine Intelligence}, 2018.

\bibitem{DBLP:journals/corr/LinMBHPRDZ14}
T.~Lin, M.~Maire, S.~J. Belongie, L.~D. Bourdev, R.~B. Girshick, J.~Hays,
  P.~Perona, D.~Ramanan, P.~Doll{\'{a}}r, and C.~L. Zitnick.
\newblock Microsoft {COCO:} common objects in context.
\newblock {\em CoRR}, abs/1405.0312, 2014.

\bibitem{M-CNN}
S.~Liu, X.~Liang, L.~Liu, X.~Shen, J.~Yang, C.~Xu, L.~Lin, X.~Cao, and S.~Yan.
\newblock {Matching-CNN Meets KNN: Quasi-Parametric Human Parsing}.
\newblock In {\em CVPR}, 2015.

\bibitem{liuLQWTcvpr16DeepFashion}
Z.~Liu, P.~Luo, S.~Qiu, X.~Wang, and X.~Tang.
\newblock Deepfashion: Powering robust clothes recognition and retrieval with
  rich annotations.
\newblock In {\em CVPR}, 2016.

\bibitem{long2014fully}
J.~Long, E.~Shelhamer, and T.~Darrell.
\newblock Fully convolutional networks for semantic segmentation.
\newblock In {\em CVPR}, 2015.

\bibitem{lu2014parsing}
W.~Lu, X.~Lian, and A.~Yuille.
\newblock Parsing semantic parts of cars using graphical models and segment
  appearance consistency.
\newblock In {\em BMVC}, 2014.

\bibitem{newell2016stacked}
A.~Newell, K.~Yang, and J.~Deng.
\newblock Stacked hourglass networks for human pose estimation.
\newblock In {\em European Conference on Computer Vision}, pages 483--499.
  Springer, 2016.

\bibitem{ouyang2014multi}
W.~Ouyang, X.~Chu, and X.~Wang.
\newblock Multi-source deep learning for human pose estimation.
\newblock In {\em Proceedings of the IEEE Conference on Computer Vision and
  Pattern Recognition}, pages 2329--2336, 2014.

\bibitem{park2017attribute}
S.~Park, X.~Nie, and S.-C. Zhu.
\newblock Attribute and-or grammar for joint parsing of human pose, parts and
  attributes.
\newblock {\em IEEE Transactions on Pattern Analysis and Machine Intelligence},
  2017.

\bibitem{Pfister15a}
T.~Pfister, J.~Charles, and A.~Zisserman.
\newblock Flowing convnets for human pose estimation in videos.
\newblock In {\em ICCV}, 2015.

\bibitem{SimoSerraACCV2014}
E.~Simo-Serra, S.~Fidler, F.~Moreno-Noguer, and R.~Urtasun.
\newblock {A High Performance CRF Model for Clothes Parsing}.
\newblock In {\em ACCV}, 2014.

\bibitem{simo2015neuroaesthetics}
E.~Simo-Serra, S.~Fidler, F.~Moreno-Noguer, and R.~Urtasun.
\newblock Neuroaesthetics in fashion: Modeling the perception of
  fashionability.
\newblock In {\em Proceedings of the IEEE Conference on Computer Vision and
  Pattern Recognition}, pages 869--877, 2015.

\bibitem{taskar2013modec}
B.~Taskar et~al.
\newblock Modec: Multimodal decomposable models for human pose estimation.
\newblock In {\em CVPR}, 2013.

\bibitem{deeppose13}
A.~Toshev and C.~Szegedy.
\newblock Deeppose: Human pose estimation via deep neural networks.
\newblock In {\em CVPR}, 2014.

\bibitem{wang2014semantic}
J.~Wang and A.~Yuille.
\newblock Semantic part segmentation using compositional model combining shape
  and appearance.
\newblock In {\em CVPR}, 2015.

\bibitem{wang2014deformable}
L.~Wang, X.~Ji, Q.~Deng, and M.~Jia.
\newblock Deformable part model based multiple pedestrian detection for video
  surveillance in crowded scenes.
\newblock In {\em Computer Vision Theory and Applications (VISAPP), 2014
  International Conference on}, volume~2, pages 599--604. IEEE, 2014.

\bibitem{wang2015joint}
P.~Wang, X.~Shen, Z.~Lin, S.~Cohen, B.~Price, and A.~Yuille.
\newblock Joint object and part segmentation using deep learned potentials.
\newblock In {\em ICCV}, 2015.

\bibitem{Wei_2016_CVPR}
S.-E. Wei, V.~Ramakrishna, T.~Kanade, and Y.~Sheikh.
\newblock Convolutional pose machines.
\newblock In {\em The IEEE Conference on Computer Vision and Pattern
  Recognition (CVPR)}, June 2016.

\bibitem{xia2015zoom}
F.~Xia, P.~Wang, L.-C. Chen, and A.~L. Yuille.
\newblock Zoom better to see clearer: Huamn part segmentation with auto zoom
  net.
\newblock In {\em ECCV}, 2016.

\bibitem{xia2016pose}
F.~Xia, J.~Zhu, P.~Wang, and A.~Yuille.
\newblock Pose-guided human parsing by an and/or graph using pose-context
  features.
\newblock In {\em AAAI}, 2016.

\bibitem{Yamaguchiparsing13}
K.~Yamaguchi, M.~Kiapour, and T.~Berg.
\newblock Paper doll parsing: Retrieving similar styles to parse clothing
  items.
\newblock In {\em ICCV}, 2013.

\bibitem{yamaguchi2012parsing}
K.~Yamaguchi, M.~Kiapour, L.~Ortiz, and T.~Berg.
\newblock Parsing clothing in fashion photographs.
\newblock In {\em CVPR}, 2012.

\bibitem{yang2016end}
W.~Yang, W.~Ouyang, H.~Li, and X.~Wang.
\newblock End-to-end learning of deformable mixture of parts and deep
  convolutional neural networks for human pose estimation.
\newblock In {\em CVPR}, 2016.

\bibitem{YangR_CVPR_2011}
Y.~Yang and D.~Ramanan.
\newblock Articulated pose estimation with flexible mixtures-of-parts.
\newblock In {\em Computer Vision and Pattern Recognition}, 2011.

\bibitem{zhao2013unsupervised}
R.~Zhao, W.~Ouyang, and X.~Wang.
\newblock Unsupervised salience learning for person re-identification.
\newblock In {\em CVPR}, 2013.

\bibitem{crfasrnn}
S.~Zheng, S.~Jayasumana, B.~Romera-Paredes, V.~Vineet, Z.~Su, D.~Du, C.~Huang,
  and P.~Torr.
\newblock Conditional random fields as recurrent neural networks.
\newblock In {\em ICCV}, 2015.

\end{thebibliography}
}

\begin{IEEEbiography}[{\includegraphics[width=1in,height=1.25in,clip,keepaspectratio]{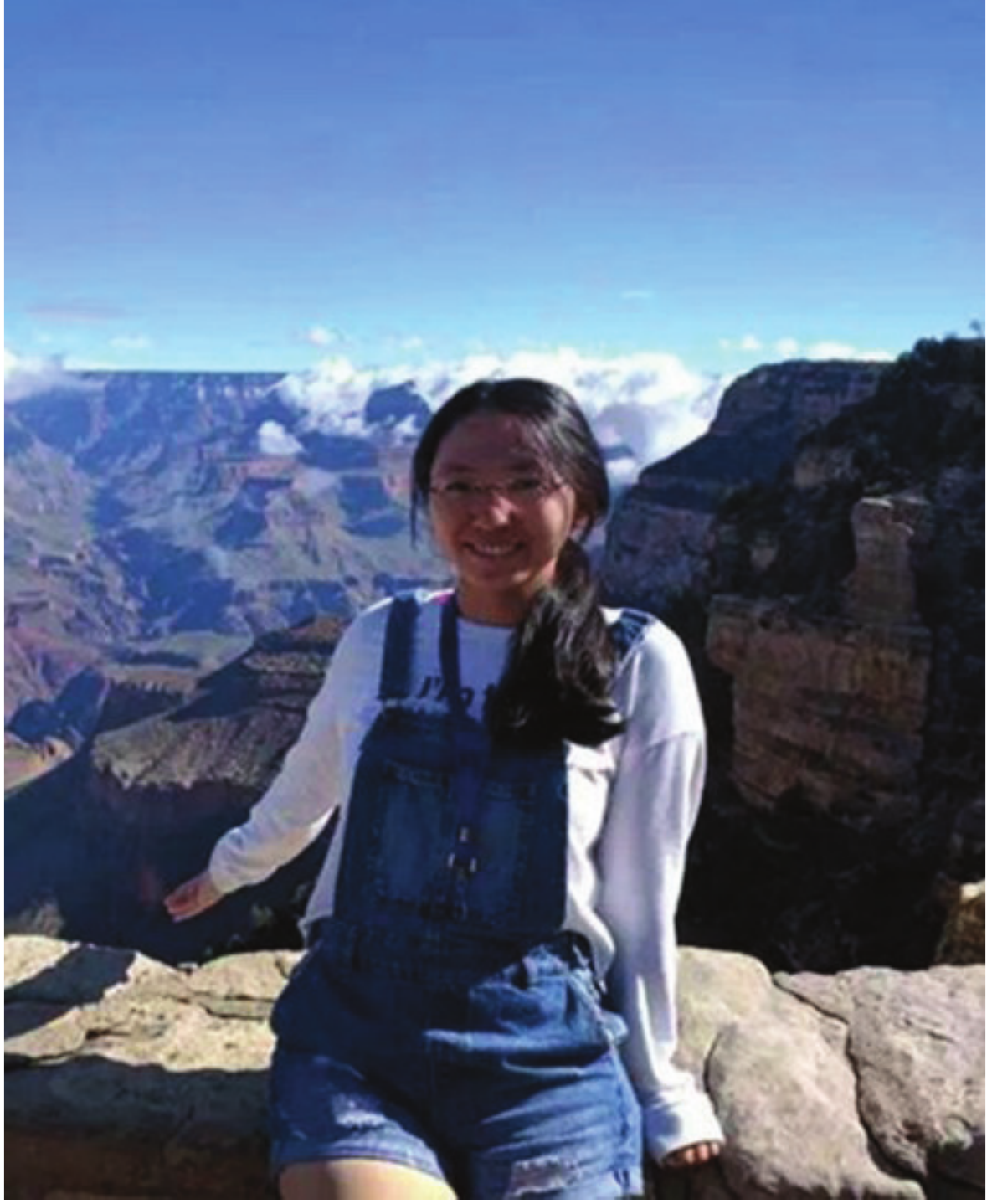}}]{Xiaodan Liang}
is currently an Associate Professor of Sun Yat-sen University. She was a postdoc researcher in Machine Learning Department at the Carnegie Mellon University, working with Prof. Eric Xing, since 2016 to 2018. She received her PhD degree from Sun Yat-sen University in 2016, advised by Liang Lin. She has published several cutting-edge projects on the human-related analysis including the human parsing, pedestrian detection and instance segmentation, 2D/3D human pose estimation and activity recognition.
\end{IEEEbiography}

\begin{IEEEbiography}[{\includegraphics[width=1in,height=1.25in,clip,keepaspectratio]{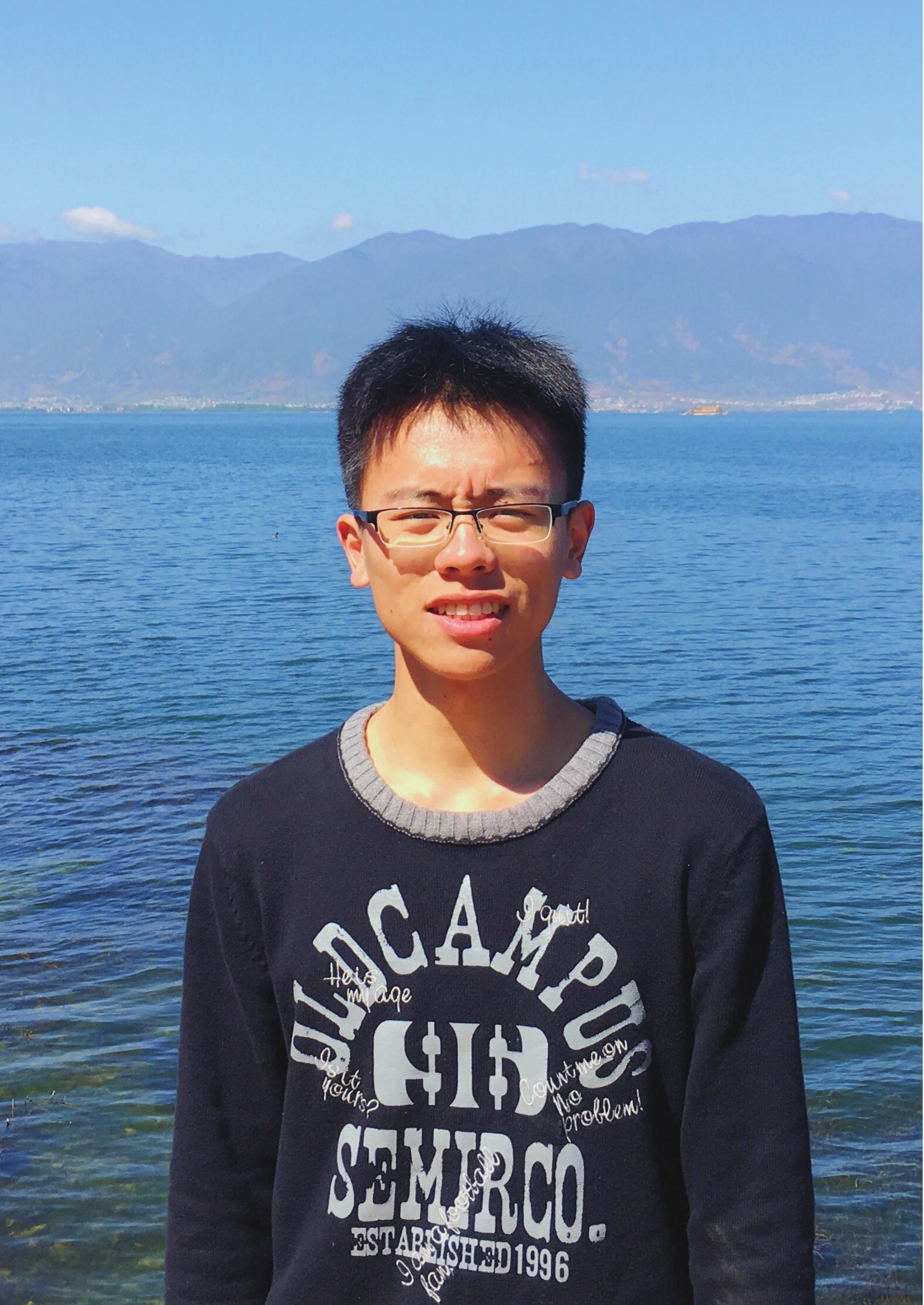}}]{Ke Gong}
received his BE degree and is currently working toward the ME degree in the School of Data and Computer Science, Sun Yat-sen University, China. His research interests include semantic segmentation and human-centric tasks, particularly human parsing and pose estimation.
\end{IEEEbiography}

\begin{IEEEbiography}[{\includegraphics[width=1in,height=1.25in,clip,keepaspectratio]{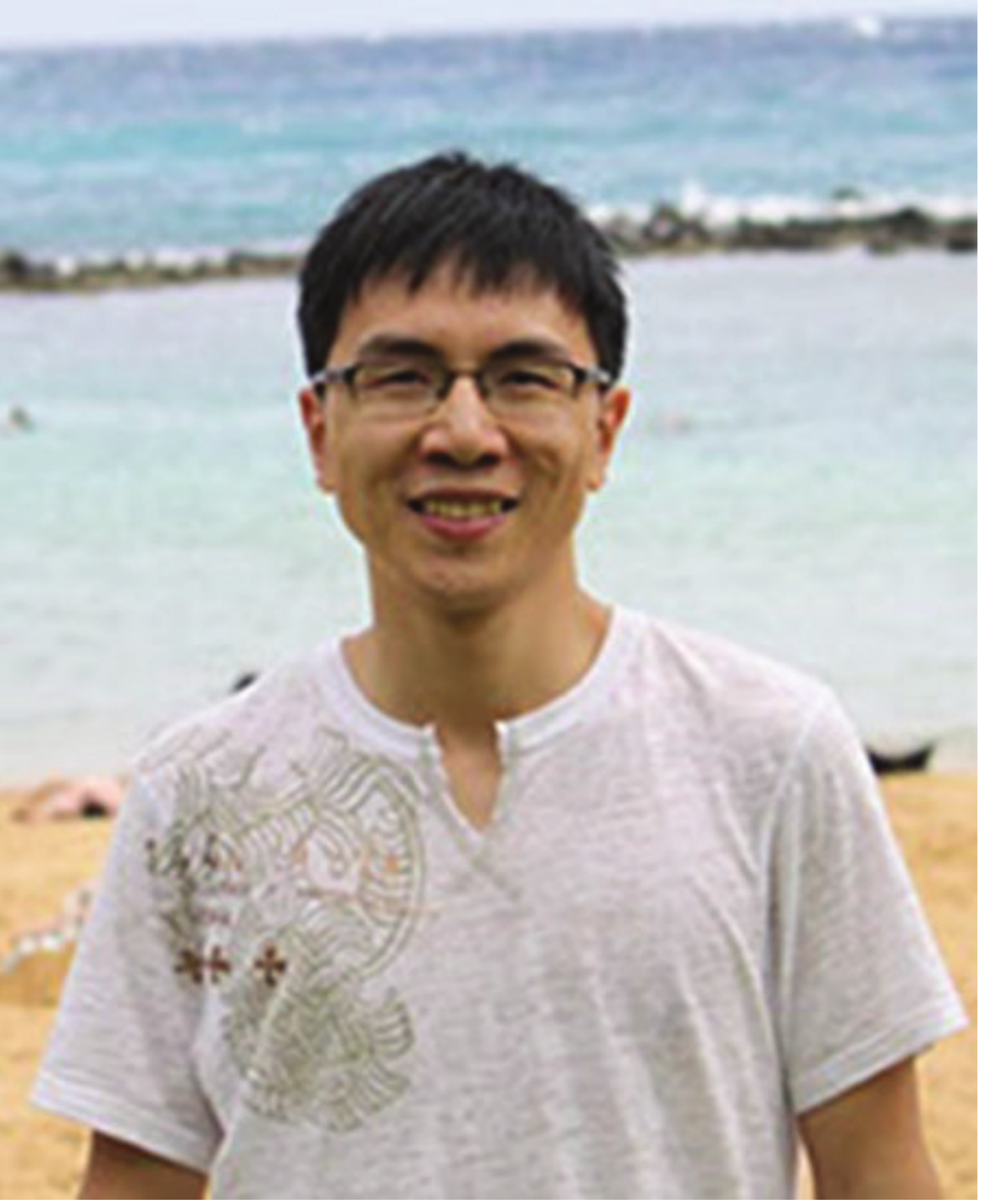}}]{Xiaohui Shen}
is a senior research scientist at Adobe Research, San Jose, CA. He received the PhD degree from the Department of Electrical Engineering and Computer Science at Northwestern University in 2013. Before that, he received the BS and MS degrees from the Department of Automation at Tsinghua University in China.  He mainly focuses on the research topics in the area of Computer Vision, particularly semantic segmentation, depth prediction, object detection and recognition, image editing and deep learning-related problems.
\end{IEEEbiography}

\begin{IEEEbiography}[{\includegraphics[width=1in,height=1.25in,clip,keepaspectratio]{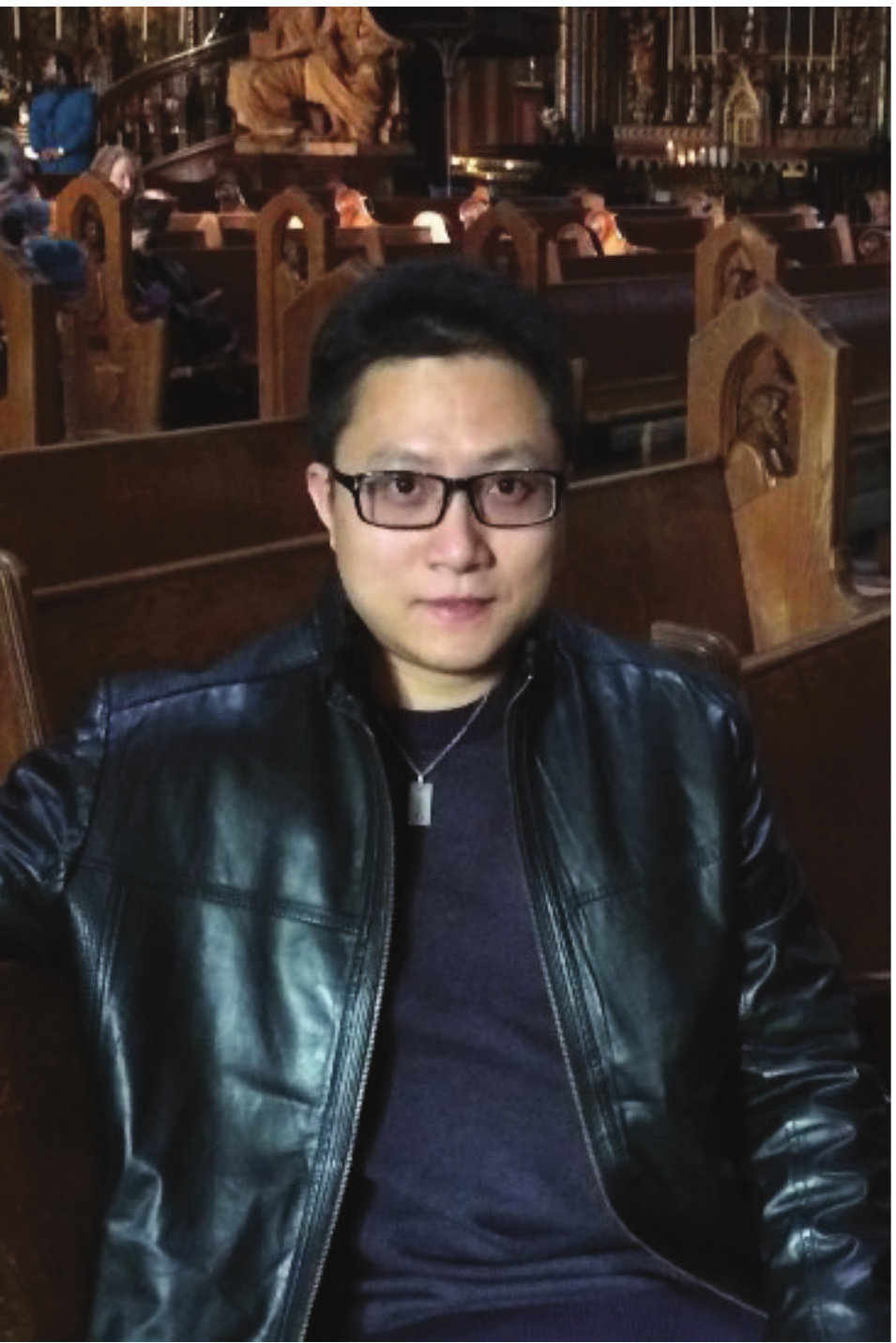}}]{Liang Lin}
(M’09, SM’15) is the Executive R\&D Director of SenseTime Group Limited and a Full Professor of Sun Yat-sen University. He is the Excellent Young Scientist of the National Natural Science Foundation of China. He has authorized and co-authorized on more than 100 papers in top-tier academic journals and conferences. He was the recipient of Best Paper Runners-Up Award in ACM NPAR 2010, Google Faculty Award in 2012, Best Paper Diamond Award in IEEE ICME 2017, and Hong Kong Scholars Award in 2014. He is a Fellow of IET.
\end{IEEEbiography}

\end{document}